\documentclass[preprint,12pt]{elsarticle}
\usepackage{graphicx}%%
\usepackage{multirow}% % Allows Table with merged cells
\usepackage{amsmath,amssymb,amsfonts}%
\usepackage{amsthm}%
\usepackage{mathrsfs}%
\usepackage[title]{appendix}%
\usepackage{xcolor}%
\usepackage{textcomp}%
\usepackage{manyfoot}%
\usepackage{booktabs}%% Allows better looking horizontal lines in tables (\toprule, \midrule, \bottomrule)
\usepackage{algorithm}%
\usepackage{algorithmicx}%
\usepackage{algpseudocode}%
\usepackage{listings}%
\usepackage{natbib}
\usepackage{parskip}
\usepackage{pslatex}
\usepackage{subcaption}
\usepackage{lineno}
\usepackage{tabularx} % For auto column width adjustments in tables
\usepackage{threeparttable} % For adding footnotes inside a table*
\usepackage{array}
\usepackage{makecell} % For manual wrap inside a cell in a table
\usepackage{url}
\usepackage{siunitx}
\usepackage[margin=2.5cm]{geometry}
\newcolumntype{L}[1]{>{\raggedright\arraybackslash}p{#1}} % Left-aligned, wrapping
\newcolumntype{C}[1]{>{\centering\arraybackslash}p{#1}}   % Centered, wrapping

\begin{document}

\begin{frontmatter}

\title{Surrogate models for Rock–Fluid Interaction: A Grid-Size-Invariant Approach \footnote{The Version of Record of this article is published in \textbf{Engineering with Computers 42, 131 (2026)}, and is available online at
https://www.doi.org/10.1007/s00366-026-02353-x}}

\author[AMCG]{Nathalie C. Pinheiro\corref{cor1}}
\ead{n.pinheiro23@imperial.ac.uk}
\author[AMCG]{Donghu Guo} %\ead{donghu.guo21@imperial.ac.uk}
\author[HWU]{Hannah P. Menke} %\ead{h.menke@hw.ac.uk}
\author[AMCG,CEE]{Aniket C. Joshi} %\ead{aniket.joshi18@imperial.ac.uk}
\author[AMCG,IX]{Claire E. Heaney\corref{cor1}} %cor2
\ead{c.heaney@imperial.ac.uk}
\author[HWU]{Ahmed H. ElSheikh} %\ead{a.elsheikh@hw.ac.uk}
\author[AMCG,IX,DAL]{Christopher C. Pain} %\ead{c.pain@imperial.ac.uk}

\address[AMCG]{{Applied Modelling and Computation Group, Department of Earth Science and Engineering}, {Imperial College London},
{{London}, {SW7 2AZ} {UK}}}
\address[HWU]{{Institute of GeoEnergy Engineering}, {Heriot-Watt University},
{{Edinburgh}, {EH14 1AS} {UK}}}
\address[CEE]{{Department of Civil and Environmental Engineering}, {Imperial College London},
{{London}, {SW7 2AZ} {UK}}}
\address[IX]{{Centre for AI-Physics Modelling, Imperial-X, White City Campus}, {Imperial College London},
{{London}, {W12 7SL} {UK}}}
\address[DAL]{{Data Assimilation Lab., Data Science Institute}, {Imperial College London},
{{London}, {SW7 2AZ} {UK}}}

\cortext[cor1]{Corresponding author}

\begin{abstract}
Modelling rock-fluid interaction requires solving a set of partial differential equations (PDEs) to predict the flow behaviour and the reactions of the fluid with the rock on the interfaces.
Conventional high-fidelity numerical models require a high resolution to obtain reliable results, resulting in huge computational expense. This restricts the applicability of these models for multi-query problems, such as uncertainty quantification and optimisation, which require running numerous scenarios.
As a cheaper alternative to high-fidelity models, this work develops 
eight surrogate models for predicting the fluid flow in porous media. Four of these are reduced-order models (ROM) based on one neural network for compression and another for prediction. The other four are single neural networks with the property of grid-size invariance; a term which we use to refer to image-to-image models that are capable of inferring on computational domains that are larger than those used during training. In addition to the novel grid-size-invariant framework for surrogate models, we compare the predictive performance of UNet and UNet++ architectures, and demonstrate that UNet++ outperforms UNet for surrogate models. Furthermore, we show that the grid-size-invariant approach is a reliable way to reduce memory consumption during training, resulting in good correlation between predicted and ground-truth values and outperforming the ROMs analysed.
The application analysed is particularly challenging because fluid-induced rock dissolution results in a non-static solid field and, consequently, it cannot be used to help in adjustments of the future prediction. 
\end{abstract}


\end{frontmatter}

%%%%%%%%%%%%%%%%%%%%%%%%%%%%%%%%%%%%%%%%%%%%
\section{Introduction}\label{secIntro}

% Background to the application
The concentration of carbon dioxide, $\mathrm{CO_2}$, in our atmosphere has increased since the mid-20th century, especially in recent decades. The impact of $\mathrm{CO_2}$ on the greenhouse effect is well-known and an increase in the global temperature can be observed in the same period~\cite{IPCC_AR6_2021}. Attempts at mitigating this sharp increase can be seen through investments in renewable energy and in Carbon Dioxide Removal (CDR) or Carbon Capture and Storage (CCS). The first refers to human activities to remove $\mathrm{CO_2}$ from the atmosphere and durably store it in reservoirs or in products~\cite{IPCC_AR6_WG3_TS_2022}. In contrast, CCS considers $\mathrm{CO_2}$ removal at the source of emissions to storage in reservoirs, avoiding additional emissions. Both CDR and CCS are crucial to limit global warming and many projects have been studied or implemented in recent years~\cite{IOGP_Map_EU_CCUS_2024}. Designing storage of $\mathrm{CO_2}$ in geological reservoirs comprises numerous studies, such as predicting plume migration within the reservoir and the dissolution of rocks induced by chemical reactions with $\mathrm{CO_2}$. 

% Motivation for the research we will present
As described in~\cite{RappMicrofluidicsCh29}, high-resolution computational tools based on CFD require significant computational resources and generate vast amounts of data. Although they have seen great improvements due to computational evolution over the last decades, using these algorithms to obtain a high accuracy demands a refined mesh and, hence, a massive computational effort. As a result, researchers have recognised an opportunity to enhance the CFD simulations by combining them with machine learning techniques, as described in~\cite{Runchal2020}. 
The main goal of this research is to develop a surrogate model framework for $\mathrm{CO_2}$ injection, which can be extended to large datasets. In fact, a key motivation for the current work is the ability of the methods to adapt to datasets with spatial domains modelled with many degrees of freedom, for example highly resolved 3D models. The surrogate models developed here can be broadly classified into two categories: (1)~reduced-order models (ROMs) which combine one neural network for compression with another network for prediction and (2)~a single neural network which has a grid-size-invariance property that allows inference over larger unseen domains.

% State-of-the-art in surrogate modelling (compression) 
ROMs typically use two neural networks, one to compress the data from physical space to a lower-dimensional latent space and another to evolve the latent variables in time. An autoencoder (AE) is well suited for compression, as it learns an identity map whilst having a bottleneck at its centre, forcing it to learn a reduced representation of the data~\cite{Lee2019model,nikolopoulos2021nonintrusive}. 
A key concept in machine learning is the convolution operation which computes element-wise products between a defined small grid of parameters~\cite{Yamashita2018_CNN}, called a filter, and the data. The purpose of this operation is to capture spatial features contained within the data~\cite{Zafar2018}. The combination of convolutional layers with AEs meant that Convolutional Autoencoders (CAEs) could be applied to larger images and that deeper CAEs could be trained successfully, as convolutional layers typically have fewer trainable parameters than networks comprising purely of fully connected layers.  One of the first reports of a CAE being used for dimensionality reduction in physical modelling occurs in \citet{gonzalez2018deep}. Since then, convolutional autoencoders have been widely applied for dimensionality reduction due to their excellent performance in extracting the main features of the original dataset. Many studies have compared its performance with other algorithms used for finding low-dimensional spaces to approximate the high-fidelity CFD models, such as POD (Proper Orthogonal Decomposition). In general, the CAEs outperform POD~\cite{maulik2021reduced, KADEETHUM2022, Fresca2021}, especially for advection-dominated flows~\cite{Heaney2022AIbased}. 

% State-of-the-art in surrogate modelling (prediction) 
For the predicion network of a ROM, the UNet has been reported to perform well. 
Primarily used for image segmentation, the UNet has been successfully applied in many other image-based applications, such as predicting wind distribution~\cite{WANG2025_UNet-WindDist}, and surrogate modelling of fluid flow, e.g., with applications including flow past a cylinder and airfoil~\cite{Le2021_SurrogCylinder}, turbulence~\cite{PANG2024_Turbulence}, and flow in porous media~\cite{Zhao2024_UNetPorousMedia, Aljamou2023_UNet, JIANG2021_UNetPorousMedia, YANG2024_PorousMediaReview, WEN2021_CO2plume,Kamrava2021}. We compare the performance of the UNet with an improved network called UNet++, a nested UNet architecture presented first in~\cite{zhou2018unetplusplus} and detailed in~\cite{zhou2019unetplusplus} for image segmentation in medical applications. Currently, UNet++ has proved to outperform the traditional UNet in various applications, such as segmentation in satellite images~\cite{Alexakis2020_UNetppSatellite}, fracture detection~\cite{Park2024_UNetppFracture}, and the creation of sketch images when combined with GANs~\cite{Abbasi2024_Pix2Pixplusplus}.

% Issues for sota surrogates and what we will investigate
A major challenge for ROMs is to produce stable and physically meaningful predictions over long periods of time. A pure data-driven surrogate model often predicts well for shorter periods of time but diverges from expected values for long-term predictions. Although the reconstruction quality of images with autoencoders can be excellent, predicting the evolution of latent variables in time often results in a loss of accuracy after a few timesteps. One reason is that the latent space is not regularised, and small variations can lead to very different reconstructions. Adversarial autoencoders~\cite{Makhzani2016_AAE} are a reasonable alternative, as they provide a regularisation of the latent space. We will compare the performance of an Autoencoder with traditional training and an Adversarial Autoencoder for compression. 
Researchers have worked on different methods to improve  prediction accuracy over long time periods, including learning operators rather than discretisations (neural differential equations~\cite{chen2019_neuralODEs, kidger2022neuraldifeqs}), physics-informed approaches~\cite{Raissi2019PINNs, CHEN2021_PRNN, Cai2021PINNs, ARTHURS2021_PINNs_N-S} and predicting for all time levels simultaneously~\cite{WEN2022_U-FNO}.
For models that are used to march forward in time, also known as autoregressive surrogates, we can find in the literature approaches that consider some rollout of predictions during training to improve how the temporal evolution is captured. 
This approach~\cite{Lusch2018_rollout,kohl2024benchmark,Nayak2025_rollout} is also referred to as unrolled training and, by minimising the modified loss, encourages the surrogate's predictions to be close to the physics-based model for a number of time steps. Here, we adopted a similar strategy for training that we call rollout training, described in Section~\ref{subsecMet-Rollout}. 

% Our grid-size-invariant approach
In addition to investigating whether adversarial training or rollout training approaches can help stabilise predictions or whether the UNet++ offers advantages in this regard, we also propose a novel grid-size-invariant approach for image-to-image neural networks. To achieve grid-size invariance, we use a particular architecture for our neural networks which enables the computational domain for prediction to be larger than the spatial extent of the samples used in training. Many of the CNNs cited above have a mixture of convolutional and fully-connected layers. However, if all the layers are convolutional, which we refer to as a fully convolutional neural network, then such a network can be applied to images of different sizes since all the connections are local~\citep{long2015FullyCNN}. The research presented here leverages this property to train the network on smaller domains while inferring on larger domains. Grid-size invariance is particularly important given that memory resources are a limiting factor in many applications, especially during training. 
Recent developments in this area include neural operators, which are point-based and therefore independent of any underlying grid~\cite{Li2020_FNO,Kovachki2023_NeuralOp,zhang2022_belnet}, whereas the proposed grid-size-invariant framework applies to grid-based and image-to-image networks. Furthermore, neural operators focus on obtaining a grid-independent model to work with any resolution, rather than being more memory-efficient. On the other hand, patch-based CNNs have been proposed for problems dealing with large images, but the network performs both training and inference on the same patch-size~\cite{OrhanBastanlar2018_CNNpatch, SHARMA2017_patch-basedCNN, goodwinallcock2023_patchCNN}. They are memory-efficient for problems such as object classification or localisation, where the input image can be divided into patches, the patches are processed individually, and then the results are combined. However, this strategy cannot be used in surrogate models where you want to predict the underlying physics. An approach similar to ours was proposed by \citet{Owerko2024_FCNs}, but in the context of mobile infrastructure on demand.

% summary of contribution
This article proposes a number of surrogate models for carbon storage in porous media and compares the model outputs to determine which methods produce the best results. The surrogate models developed here fall into two categories: (1)~ROMs and (2)~grid-size-invariant neural networks. For compression in the ROMs, adversarial training is investigated. For the prediction models (both in the ROMs and the grid-size-invariant networks), the performance of UNet and UNet++ is investigated. The surrogate models are trained on a 2D carbon storage dataset~\citep{Maes2022improved}. Although this dataset contains 2D-images with $256 \times 256$ pixels and represents flow in a porous medium, the proposed framework is quite general and could be applied to other fluid flow applications as well as larger datasets that could come from highly resolved CFD simulations. For example, a similar framework with implicit time marching is being developed for inertial flows~\citep{Guo2025}. 

Besides the development of a grid-size invariant surrogate model for carbon injection, a second contribution of this work is an extensive investigation of eight surrogate models, highlighting performance improvements when changing the framework, architecture or training strategy. Furthermore, a particular challenge of the carbon storage application analysed here is that the dissolution of the rock caused by the chemical reactions with the fluid results in a non-static solid field. Consequently, it cannot be used as a special feature capable of helping in adjusting future predictions, such as some other surrogate models, which use the solid field as a mask for correcting predictions~\cite{WANG2025_InpaintingStaticField,HEMMASIAN2023_mask,zhou2022_maskedsurrog}. 

In the remainder of the text, we explain the methodology of the proposed models in Section~\ref{secMet}; present the results in Section~\ref{secResults}; and draw conclusions and discuss future work in Section~\ref{secConclusions}.

%%%%%%%%%%%%%%%%%%%%%%%%%%%%%%%
\section{Methodology}\label{secMet}

This work compares different strategies for surrogate models. Four reduced-order surrogate models are developed, which couple an autoencoder for compression with another neural network for prediction. For compression, we compare traditional and adversarial training. For prediction, we compare UNet and UNet++ architectures. These reduced-order models are compared with four single neural networks that have grid-size-invariant properties,
which enables inference over larger (unseen) domains.
Table ~\ref{tab:Models_summary} summarises the characteristics of each model, and the following sections explain them in detail.

\begin{table*}[h]
\centering

\begin{threeparttable}
\caption{Comparison of model architectures and training strategies. }
\label{tab:Models_summary}

\begin{tabular*}{\textwidth}{@{\extracolsep\fill}llcccc}
\toprule

\makecell[c]{} & 
\makecell[c]{\textbf{Method}} & 
\makecell[c]{\textbf{Total} \\ \textbf{Parameters}\footnotemark[1]} &  
\makecell[c]{\textbf{Compression} \\ \textbf{Architecture}\footnotemark[2]} & 
\makecell[c]{\textbf{Prediction} \\ \textbf{Architecture}} & 
\makecell[c]{\textbf{Training}} \\ 
\midrule
\multirow{4}{*}{ROMs}
%\multirow{4}{*}{\makecell[l]{ROMs \\ (C)+(P)}} 
 & AE + UNet & 855k + 7.7M & Enc-Dec\_4HL & UNet (depth=4) & Traditional \\
 & AE + UNet++ & 855k + 9M  & Enc-Dec\_4HL & UNet++ (depth=4) & Traditional \\
 & AAE + UNet & 856k + 7.7M & Enc-Dec\_4HL & UNet (depth=4) & Adversarial \\
 & AAE + UNet++ & 856k + 9M & Enc-Dec\_4HL & UNet++ (depth=4) & Adversarial \\ 
\midrule
\multirow{4}{*}{\makecell[l]{Grid-Size \\Invariant}} 
 & UNet & 7.7M & - & UNet (depth=4) & Traditional \\
 & UNet rollT8 & 7.7M & - & UNet (depth=4) & Rollout \\
 & UNet++ & 9M   & - & UNet++ (depth=4) & Traditional \\
 & UNet++ rollT8 & 9M  & - & UNet++ (depth=4) & Rollout \\ 
\bottomrule
\end{tabular*}

\begin{tablenotes}
%\footnotesize
\item[1] For ROM models, information presented as C + P, where C is for the compression network and P for the prediction network.
\item[2] ``Enc-Dec\_4HL'' indicates an Encoder-Decoder architecture with 4 hidden layers in encoder and in decoder. The number of neurons in each hidden layer is: 40; 200; 200; 40.  For the AAE, the Encoder-Decoder has the same configuration, and the discriminator used for training comprises two convolutional layers.
\end{tablenotes}

\end{threeparttable}

\end{table*}

\subsection{Reduced-order models}\label{subsecMet-ROM}
\subsubsection{Convolutional autoencoder for compression} \label{subsec-CAE} %\comment{Shortened version}

An autoencoder mimics the identity map and comprises two parts either side of a bottleneck~\cite{Phillips2021_AE}. This structure can be used to force the network to learn a compressed representation of the input. The first part, referred to as an encoder, compresses the input into a latent-space representation, where the size of the latent space ($m$) is much less that the size of the physical space or input ($n$), i.e., $m \ll n$. The second part, the decoder, reconstructs the latent space to the physical space and generates the final output of the autoencoder~\cite{Lee2019model}. 

For a neural network with hidden layers, each layer corresponds to a linear transformation of the input $x$ composed with an activation function to account for the non-linearities. Fully connected layers are such that a neuron is connected to every neuron in the preceding layer, whereas convolutional layers connect a neuron to only a small subset of neurons in the preceding layer. This is done by applying a filter of kernel size $K$ (a small array of weights with size $K \times K$) through the input matrix, and moving the filter by a certain number of pixels in each dimension, number known as a stride $s$~\cite{nikolopoulos2021nonintrusive}. 

When training an autoencoder, a gradient-based optimisation method is applied to minimise the loss between the output of the network and the original input, thereby determining the weights of the neural network. In the results presented in this work, we used mean-square error as the loss function, given by:
\begin{equation}
\mathcal{L}_{\text{MSE}} = \frac{1}{n} \left\| \mathbf{X} - \tilde{\mathbf{X}} \right\|_F^2 
=  \frac{1}{n} \sum_{k=1}^{n} \left( x_k - \tilde{x}_k \right)^2,
\label{eq_MSE}
\end{equation} 

where $n$ is the number of entries in the input matrix $\mathbf{X}$, $\tilde{\mathbf{X}}$ is the output of the autoencoder and subscript $F$ denotes the Frobenius norm.

\subsubsection{Adversarial training}\label{subsec-Adversarial}

In this research, we compare the traditional training of an Autoencoder (AE) with an adversarial training forming an Adversarial Autoencoder (AAE). This strategy for training was first proposed for Generative Adversarial Networks (GANs)~\cite{Goodfellow2014_GAN}.
During adversarial training, one step involves updating the discriminator weights based on the loss when trying to distinguish between fake and real inputs. The other step is updating generator weights to fool the discriminator. In this case, the idea behind calculating the loss function is to maximise the probability of the discriminator making a mistake. This algorithm consists of a minimax two-player game~\cite{Goodfellow2014_GAN}, and the loss is given by the cross-entropy function:

\begin{equation}
\min_{G} \max_{D} \; \mathcal{L}(D, G) 
= \mathbb{E}_{\mathbf{x} \sim p_{x}} \big[ \log D(\mathbf{x}) \big] 
+ \mathbb{E}_{\mathbf{z} \sim p_{\mathbf{z}}} \big[ \log (1 - D(G(\mathbf{z}))) \big]. \label{eq_minmax}
\end{equation}

For Adversarial Autoencoders, the training process aims to impose a specific prior distribution on the latent space of the autoencoder. In this case, the generator is the encoder, and the discriminator should compare its result with a sample with the dimensions of the latent space and with the prior distribution, $p_x$, usually a normal distribution. The training for an AAE has an additional step, where the weights of the encoder and decoder are updated to minimise the loss in reconstruction, according to equation~\eqref{eq_MSE}. Then, the discriminator is updated, followed by the encoder acting as a generator, considering the minimax loss in equation~\eqref{eq_minmax}. 
A more detailed mathematical description of the loss function and the backpropagation for the adversarial autoencoder can be found in~\cite{ghojogh2021generative}.

%%%%%%%%%%%%%%%%%
\subsection{UNet and UNet++ for prediction}\label{subsecMet-UNetvsUNetpp}
\begin{figure}[!htb]
	\centering
	\includegraphics[width=\textwidth]{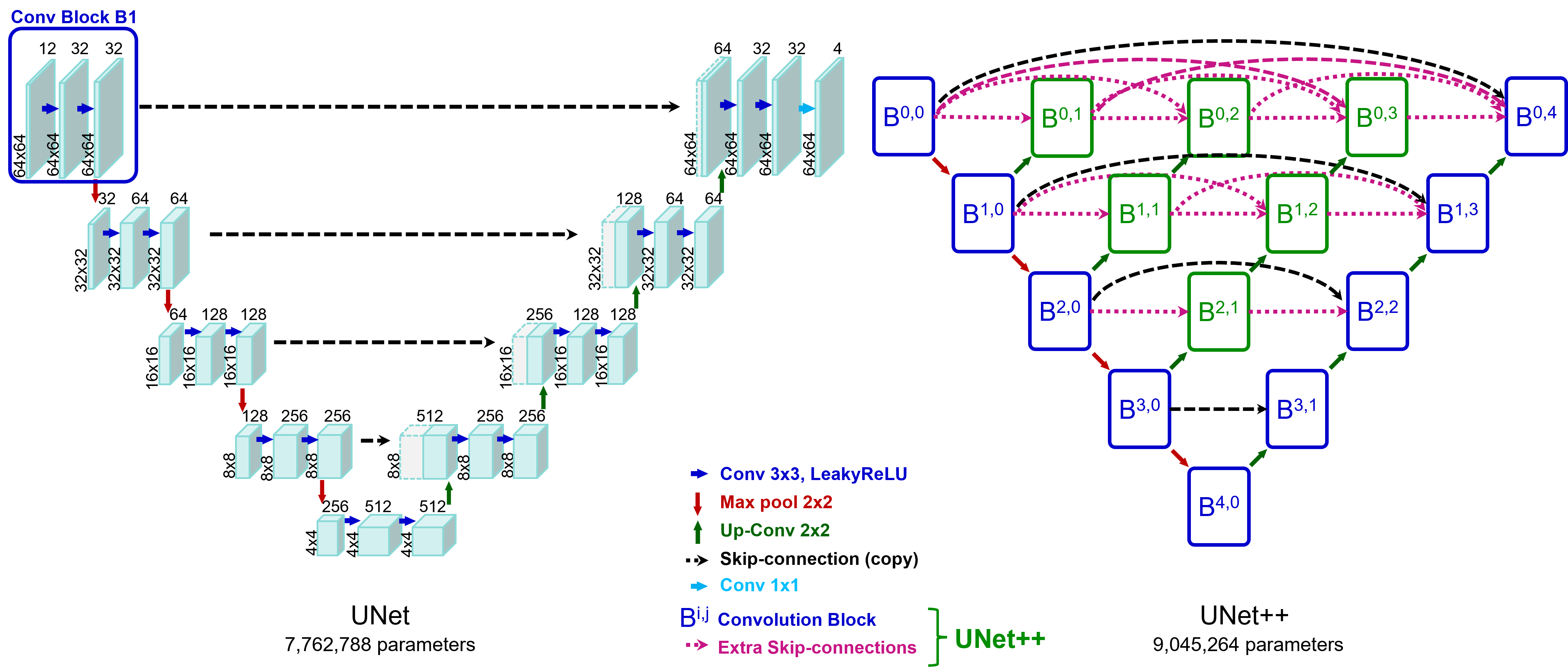}
	\caption{Diagrams for the two different options of the prediction neural network compared. In the UNet++ diagram, each block $B^{i,j}$ corresponds to the sequence of operations highlighted in ``Conv Block B1'' in the UNet diagram.}
	\label{fig:Pred_UNetvsUNetpp}
\end{figure}

To predict the spatio-temporal evolution of the latent variables, we compare a UNet and a more complex network derived from it, the UNet++, both with four downsampling levels. Figure~\ref{fig:Pred_UNetvsUNetpp} illustrates their architectures, and we can observe that UNet++ has additional convolutional blocks, creating extra connections between encoder and decoder layers. The hypothesis behind this architecture is that the extra blocks bring the semantic level of the encoder feature maps closer to that awaiting in decoder, hence capturing fine-grained details~\cite{zhou2018unetplusplus}. In both UNet and UNet++, the network input consisted of three previous timesteps stacked along the field dimension, and the predictor generated one next timestep. Thus, the total input channels is 12 (3 timesteps for 4 fields) and the number of output channels is 4. The diagram for UNet details the number of channels in each layer above the blocks, and shows how the spatial dimensions change in the vertical text next to each layer, considering training with $64x64$ samples. For the UNet++ diagram, this numbers were omitted, but follows a similar logic: you have an encoder backbone (blocks from $B^{0,0}$ to $B^{4,0}$) where each block comprises one convolution increasing the number of channels and one convolution maintaining the number of channels; and the other blocks comprises one convolution decreasing the channels and a second convolution maintaining it. As in the UNet, the UNet++ also maintain the spatial dimensions in these convolutions, only changing it through max pool operations (red arrows) or up-conv (green arrows).

%%%%%%%%%%%%

\subsection{Grid-size-invariant framework}\label{subsecMet-GridInv}

The grid-size-invariant framework (GSI) relies on the ability of fully convolutional networks to capture spatial features and their applicability to arrays of any size due to the connections being local~\cite{long2015FullyCNN}. Considering these properties, a convolutional-based network can be trained with smaller domains and used for inference over larger unseen domains. Thus, we refer to the ability of the neural network to be applied to a domain of any size as the grid-size-invariance property. Nevertheless, the grid spacing 
must be the same as that used to generate the training data. 

\begin{figure}[!htb]
	\centering
	\includegraphics[width=0.5\textwidth]{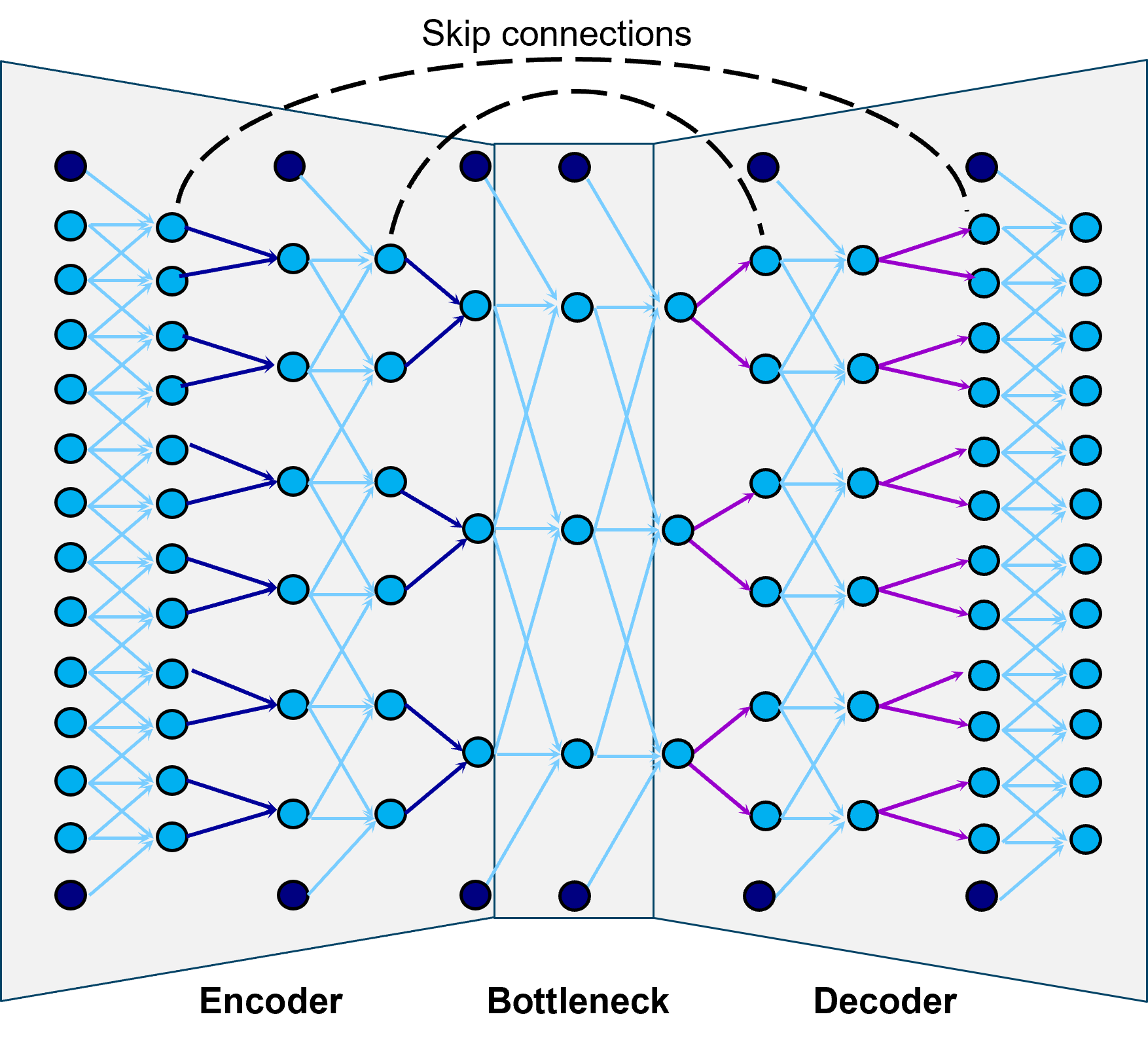}
	\caption{Grid-size-invariant architecture of a simplified UNet (light blue arrows in smoothing layers, dark blue in reducing layers, and purple in prolongation layers.)}
	\label{fig:grid-inv-diagram}
\end{figure}

Besides using only CNNs, the right choice of kernel size and stride in each layer helps achieve good performance with grid-size invariance. For example, Figure~\ref{fig:grid-inv-diagram} illustrates the connections in a GSI convolutional network for one-dimensional data. The first layer (from the left) is a smoothing layer with kernel size~3 and stride~1 (dark nodes indicate padding, which is necessary when using a kernel size of 3). The next layer is a downsampling layer with a kernel size of~2 and a stride of~2. There follows another smoothing layer, another downsampling layer and a final smoothing layer. Having a stride greater than~1 separates the data into regions in a manner similar to domain decomposition. The smoothing layers facilitate communication between these regions, enabling changes in the number of channels to perform domain reductions more smoothly (for simplicity, the diagram does not show the channels, which can be understood as parallel nodes for each representation of a node in the diagram). The decoder part of the network has a similar structure, however the reducing layers are substituted by prolongation layers, where the domain is doubled. It is this structure which gives the networks their grid-invariant property.

A grid-invariant architecture can be built into different types of neural network. Here, we compare UNet and UNet++ architectures with the same structure shown in Figure~\ref{fig:Pred_UNetvsUNetpp}, both based on convolutional layers, and rely on the grid-size-invariance property for subsampling for training.

It is noteworthy that the training subsamples need to be representative of the geometries and behaviour considered. For instance, if the majority of the fluid flows in a free space and at some point it passes through an obstacle, we need to ensure that our training samples represent both flow through free space and the interaction of flow with obstacles. In the porous media flows studied in this paper, the samples used in training are larger than the pore sizes, thus containing a representative sample of physical and chemical interactions.

%%%%%%%%%%%%%%%%%%%%%%%
\subsection{Rollout training}\label{subsecMet-Rollout}

\begin{figure}[!htb] %[htbp]
\centering
\includegraphics[width=0.75\textwidth]{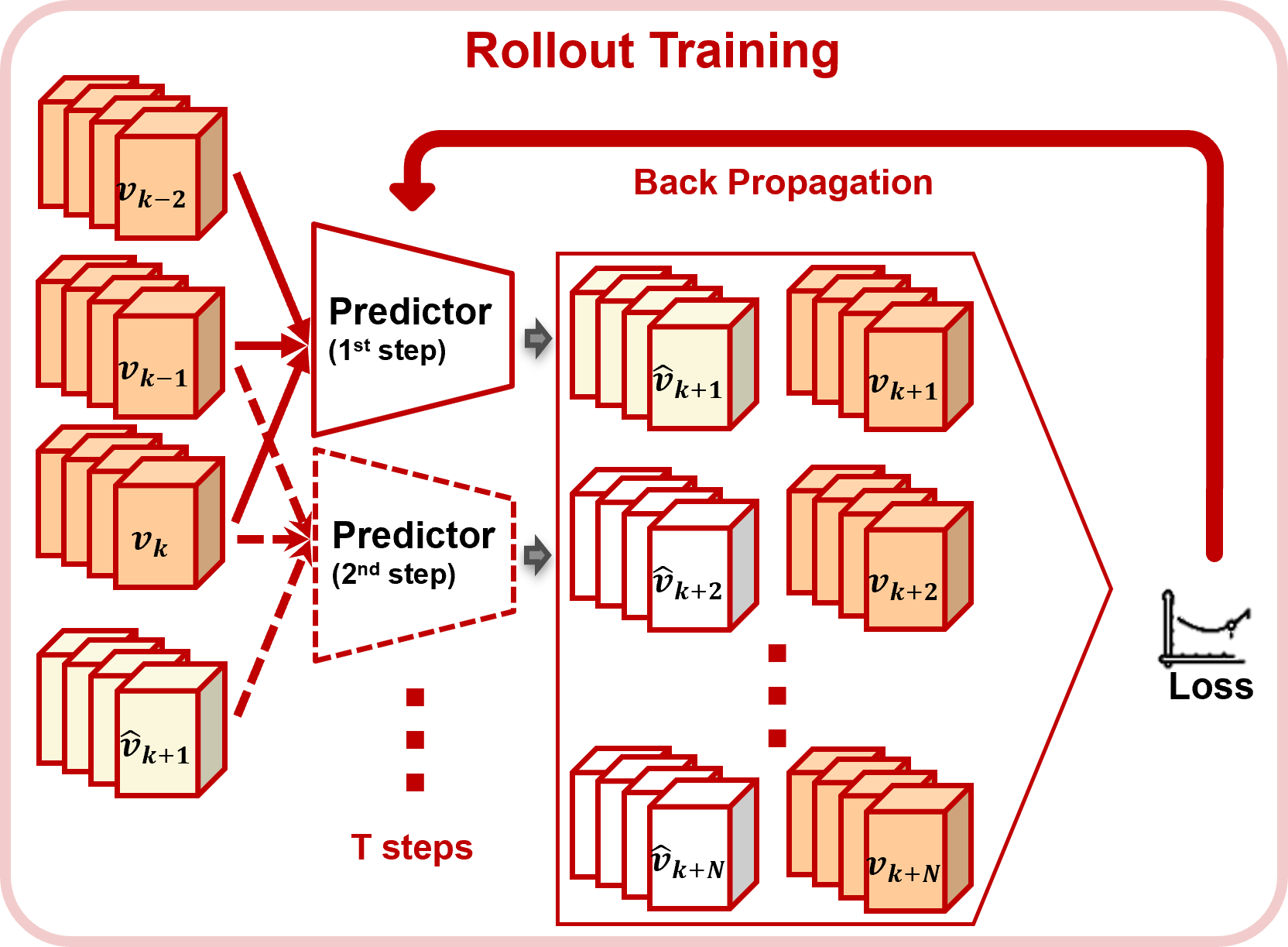}
\caption{Rollout training: strategy used to improve multiple timesteps inference in grid-size-invariant framework.}\label{fig:RolloutTrainDiagram}
\end{figure}

In the conventional approach for training, a surrogate model is trained to predict a single timestep. After that, the model is applied to predict multiple time steps autoregressively. Consequently, the following predictions carry the errors of the previous ones, accumulating these errors. As the model is trained to optimise only the next prediction, the accumulated error can increase quickly. To reduce the accumulated error in the grid-size-invariant framework, we employ rollout training. In this strategy, we iterate for a predetermined number of timesteps $T$ inside the training loop, and measure the loss between the predicted and the true values altogether for the $T$ timesteps. This is shown in the diagram of Figure~\ref{fig:RolloutTrainDiagram}. Therefore, the model is not only trained to predict the next timestep, but it should also consider the predictions over multiple timesteps. 

As the rollout training approach is more expensive in terms of memory and time consumption, we utilise the concept of curriculum learning to reduce the number of epochs required for convergence. When using curriculum learning, the network is first trained with simpler samples, and then trained with the most complex cases~\cite{Bengio2009_CurriculumLearning, soviany2022_CurriculumLearning}.
In our case, we first run a conventional training (one timestep loss), then initialise the predictor with the obtained weights to retrain and adjust the weights to account for the behaviour over $T$ timesteps. With this strategy, the rollout converges to its best result after a few epochs.

%%%%%%%%%%%%%%%%%%%%%%%
\subsection{Boundary conditions}\label{subsecMet-BC}

To enforce a boundary condition during neural network training, we used a strategy that penalises errors at the boundaries: we compute the loss during training using mean squared error (MSE) in the whole spatial domain for prediction $\Omega$, and add an extra term with a multiplier $\lambda_{BC}$ and the MSE only at the boundary $\partial \Omega$. Therefore, the multiplier $\lambda_{BC}$ serves as a hyperparameter to be tuned to penalise more the loss in the boundaries.
\begin{equation}
\mathcal{L}_{\text{Total}} =\mathcal{L}_{\text{MSE}} \rvert_{\Omega} + \lambda_{BC} \cdot \mathcal{L}_{\text{MSE}} \rvert_{\partial \Omega},
\label{eq_Total_loss}
\end{equation} 
The domain $\Omega$ for the prediction in the ROM models is a $64 \times 64$ space, with images generated in latent space of compression network. For the grid-size-invariant framework, the domain $\Omega$ is the space of the original images ($256 \times 256$). The models that are presented here considered $\partial \Omega$ as the outer 1-pixel boundary of the domain $\Omega$. 
Initially $\lambda_{BC}=0$, that is, the training starts without any regularisation in the boundaries. Then, we change to $\lambda_{BC}=0.5$ and finally $\lambda_{BC}=1$, as a way of increasing the importance given to the solution on the boundaries. The time for changing the $\lambda_{BC}$ is empirical and depends on the number of epochs to reach the best model. For ROMs, the predictor was trained over $1000$ epochs, and $\lambda_{BC}$ was updated at epoch numbers~$100$ and~$200$. For the grid-size-invariant framework, the loss during training stops decreasing earlier, especially for UNet++ models. Thus, we updated $\lambda_{BC}$ at epoch numbers~$50$ and~$100$ for the UNet, and~$15$ and~$65$ for the UNet++.

%%%%%%%%%%%%%%%%%
\section{Numerical results}\label{secResults}

This section presents the results obtained from the surrogate models for a carbon storage scenario with rock-fluid interaction. Two strategies are investigated: (1)~a ROM-type approach involving compression to latent space, prediction of latent variables and reconstruction to the physical space; (2)~prediction with a single grid-size-invariant neural network.

%%%%%%%%%%%%%%%
\subsection{Governing equations and the carbon storage dataset}\label{subsecIntro-Problem}

The rock-fluid interaction problem, which appears in carbon storage injection, is governed by equations that describe fluid flow, transport, and reaction at the fluid-solid interface. Under isothermal conditions and in the absence of gravitational effects, fluid motion is governed by the incompressible Navier-Stokes equations~\cite{Maes2022improved}:
\begin{align}
\nabla \cdot \mathbf{u} &= 0, \\
\frac{\partial \mathbf{u}}{\partial t} + \nabla \cdot (\mathbf{u} \otimes \mathbf{u}) &= -\nabla p + \nu \nabla^2 \mathbf{u},
\end{align}

where $\textbf{u}$ (\si{m/s}) is the velocity, $p$ (\si{m^2/s^2}) is the kinematic pressure, and $\nu$ (\si{m^2/s}) is the kinematic viscosity.

The transport is governed by the advection-diffusion equation, which states that the concentration $c$~(\si{\kilo\mole\per\meter\cubed}) of a species in the system, subject to the diffusion coefficient $D$ (\si{m^2/s}), should satisfy~\cite{Maes2022improved}:

\begin{align}
\frac{\partial c}{\partial t} + \nabla \cdot (\mathbf{u}c) = \nabla \cdot (D \nabla c).
\end{align}

Finally, these equations are subject to continuity conditions at the fluid-solid interface based on the chemical reaction that occurs between the fluid and the solid. Using improved Volume-of-Solid (iVoS) to calculate the volume-averaged surface reaction rate $\overline{R}$ at the interface,

\begin{align}
\overline{R} = \varepsilon \nabla \cdot \overline{\Phi}_R - \nabla \cdot (\varepsilon \overline{\Phi}_R),
\end{align}

where $\varepsilon$ is the local porosity, and $\overline{\Phi}_R$ is an approximation for the reactive flux, assuming that the reactive surface can be approximated by its volume-averaged value on the control volume. 
\citet{Maes2022improved} proposed the use of iVoS method above for reactive transport modelling in porous media. A more detailed explanation about the derivation of the interface reaction term can be found in~\cite{Maes2022improved}, where the authors present a description of how they generated simulations with GeoChemFOAM. Their simulations show how $\mathrm{CO_2}$ flow evolves in various porous media and provokes rock dissolution, and were used to create a dataset used to train and validate the surrogate models presented here.

The carbon storage dataset produced by \citet{Maes2022improved} contains 2D images obtained from 32 simulations of carbon dioxide injection in a carbonate reservoir, each simulation considering a different porosity. The porosity fields were created using a micromodel that has been benchmarked with an experimental investigation described in~\cite{Dimou2022}. The simulations of $\mathrm{CO_2}$ injection were generated in a solver called GeoChemFOAM, with a grid spacing of \SI{25}{\micro\meter} in each direction. The snapshots were taken every \SI{4000}{s}, although each snapshot corresponded to thousands of numerical timesteps in the GeoChemFOAM simulation (the number varies according to the convergence criterion for outputting results). 
For each simulation, the dataset contains 100 timesteps of $256 \times 256$ images, representing four fields: concentration of $\mathrm{CO_2}$, porosity, and velocities in the $x$- and $y$-directions. 

The dataset was divided into a training dataset, which contains all the sequential images for 24 simulations, and a validation dataset, with images of the remaining 8 simulations. Therefore, all networks were trained only on images from the training dataset, and the remaining dataset was used to verify the generalisation capability of the models.

%%%%%%%%%%%%%%%%%%%%%%%%
\subsection{Metrics}\label{subsecMet_metrics}

We compare the prediction methods using the Pearson Correlation Coefficient (PCC) as a metric. This coefficient measures the linear association between two variables (in this case, prediction array $X$ and ground truth array $Y$) and has a range of $[-1, 1]$. The larger absolute values indicate a stronger relationship between the variables, with 0 indicating no relationship between the variables~\cite{StatisticsInANutshell_Boslaugh}.

\begin{align}
PCC(X,Y) =  \frac{\sum (x_i - \overline{x})(y_i - \overline{y})}{\sqrt{\sum{(x_i - \overline{x})^2}}{\sqrt{\sum{(y_i - \overline{y})^2}}}},
\end{align}

where $x_i$ and $y_i$ are the individual elements for $X$ and $Y$, and $\overline{x}$ and $\overline{y}$ are the mean values for $X$ and $Y$, respectively. We can observe that the numerator is the covariance between $X$ and $Y$, and the denominator is the product of the standard deviations of $X$ and $Y$.

In case of prediction in a compressed space, this metric could be used to analyse only the performance of the prediction network, comparing the expected and predicted values in the compressed space, or the overall performance, comparing the results after reconstruction. Here, we show the results of the PCC after reconstruction.

We also compare the grid-size-invariant frameworks through the metrics of Structural Similarity Index Measure (SSIM), which was first introduced by Wang et al in 2004 in~\cite{Wang2004_SSIM}. Since then, it has been a very popular metric for image quality, as it relies on the comparison of luminance, contrast and structure of both images, efficiently capturing human perception of the difference in images. As shown in~\cite{Wang2004_SSIM}, the similarity for each pixel in our prediction image $x_i$ and the corresponding pixel in our ground truth image $y_i$ can be expressed as

\begin{equation}
SSIM(x_i, y_i) = \frac{(2\mu_x \mu_y + C_1)(2\sigma_{xy} + C_2)}{(\mu_x^2 + \mu_y^2 + C_1)(\sigma_x^2 + \sigma_y^2 + C_2)},
\end{equation}

where $\mu_x$ and $\mu_y$ are local means, and $\sigma_x$ and $\sigma_y$ are local variances of $x_i$ and $y_i$, respectively; $\sigma_{xy}$ is a local covariance between $x_i$ and $y_i$, and $C_1$ and $C_2$ are small constants to avoid division by zero. These local statistics are computed by applying a Gaussian filter for each pixel, and then the final SSIM is calculated as the mean of all $SSIM(x_i, y_i)$ between both images.

Finally, we propose a metric tailored for this carbon storage application and use it in comparisons of grid-size-invariant frameworks. This metric measures the difference in area occupied by $\mathrm{CO_2}$ in original simulations (ground truth) and the area predicted in the proposed models. Thus, we define a threshold value in the middle of concentration range, $C_{\mathrm{threshold}}$, and compute this metric as the difference in percentage of number of pixels above this threshold in ground truth concentration field $Y_n$ and in prediction $\tilde{Y}_n$, that is,

\begin{equation}
    E_{areaCO_2} = 100 \left( 
\frac{A(Y)-A(\tilde{Y})}{N}
\right)\,,
\label{eq_areaCO2}
\end{equation}
where $A(Y)= |\{n\in\{0,1,\ldots,N\}:Y_n>C_{\mathrm{threshold}}\}|$ is the cardinality of the set of pixels whose value of concentration ($Y_n$) exceeds that of the threshold, 
and $N$ is the total number of pixels for the concentration field.

\subsection{Reduced-order model surrogates}\label{subsecR-Comp}

This framework utilises two nested neural networks: the first for compressing the input data and the second for predicting the future timestep in the latent space. As the size of the data used in the prediction network is reduced, this approach is part of a set of surrogate frameworks called Reduced Order Models (ROM).

\begin{figure}[h]
\centering
\includegraphics[width=.98\textwidth]{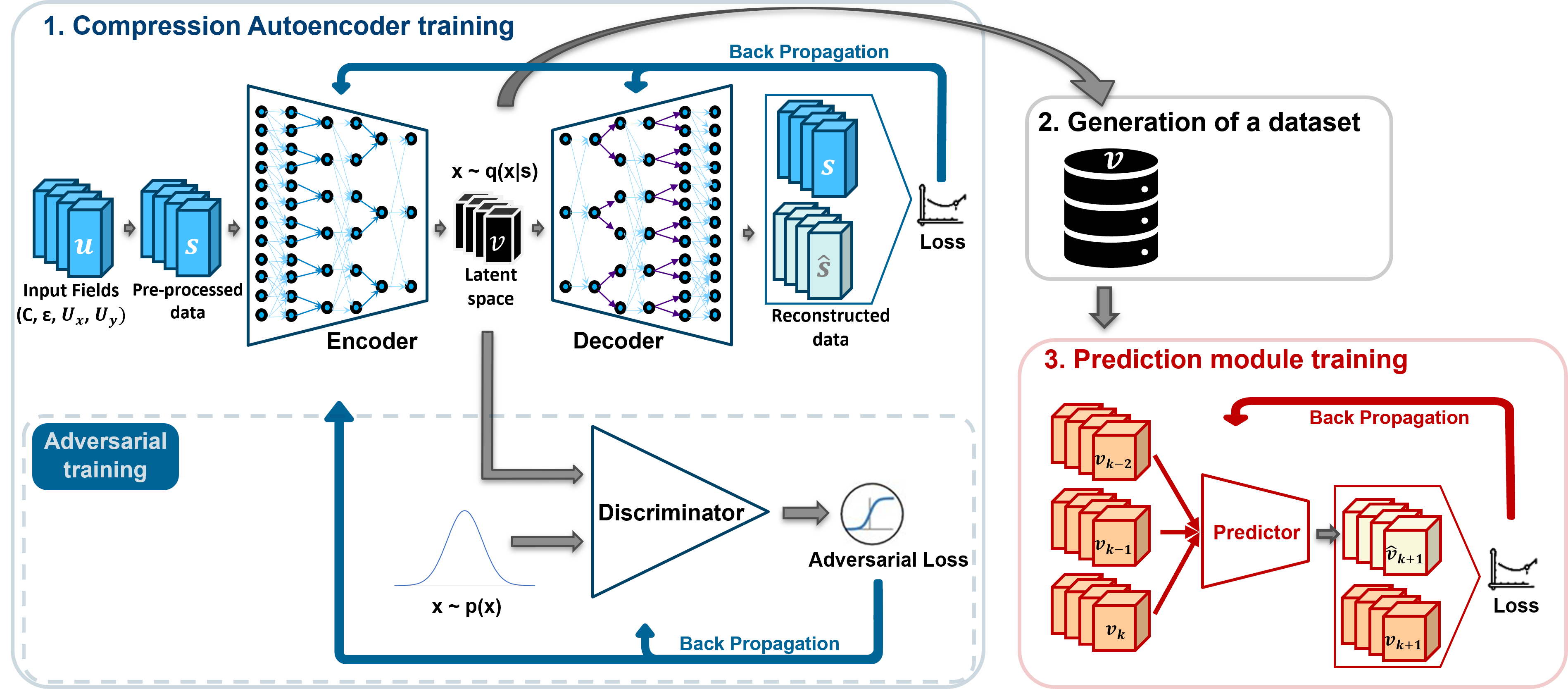}
\caption{Workflow for training the ROMs. The dashed box indicates the adversarial training extra steps, used only in Adversarial Autoencoder (AAE). After training the compression network, a prediction network is trained using latent space of compression.} \label{fig:SurrogCompressionWorkflow-Training}
\end{figure}

The first step of compression is a pre-processing step where min-max scaling is applied to each of the four solution fields to limit the values of inputs to the range $[0,1]$. The compression network is based on a convolutional autoencoder (CAE), which consists of an encoder followed by a decoder. In our case, the encoder has two layers which reduce the size of the image, with connecting layers in between. The autoencoder (AE) is trained to minimise the difference between the reconstructed data (decoder output) and the original input data. The backpropagation procedure determines the weights of the neural network, finding values that minimise the error at each epoch of training. See Figure~\ref{fig:SurrogCompressionWorkflow-Training} for a schematic diagram of the training process. After successful training of the AE, it has weights which result in the reconstructed data being similar to the original input data. In our investigation, we train two autoencoders, one as described above and a second with an adversarial autoencoder training approach (AAE) described in Section~\ref{subsec-Adversarial}.

We now present results comparing the behaviour of the AE and AAE for compression. Both networks reduce the size of the 2D solution fields by a factor of 4 in each dimension, resulting in a memory reduction ratio of $16:1$. The reconstructed results of the AE and AAE applied to a validation sample are nearly indistinguishable from the original fields, as shown in Appendix \ref{ap:compression}, where the details about the hyperparameter tuning and results are presented. The Mean Squared Error (MSE) obtained for the reconstruction of training data was \num{5.7e-6}, and in validation data was \num{1.1e-5} for the autoencoder. For the adversarial autoencoder, the MSE value was \num{3.8e-5} in training data and \num{8.7e-5} in validation data.

After training the AE and AAE, the encoders are ready to generate datasets that will be used to train neural network for prediction. The encoder of the AE finishes with a sigmoid activation function, which ensures the output is in $[0,1]$. The AAE has no activation function at the end of the encoder, but the adversarial training regularises the latent space and prevents the values from rising excessively. Thus, no scaling of the latent variables from the autoencoders is needed before this data is used to train the prediction networks. This work compares some different networks for prediction, but always considers the four solution fields at three consecutive timesteps as the input and those four fields at the future timestep as the output. The training of the entire framework developed is represented in Figure~\ref{fig:SurrogCompressionWorkflow-Training}. 

\begin{figure}[!h]
\centering
\includegraphics[width=\textwidth]{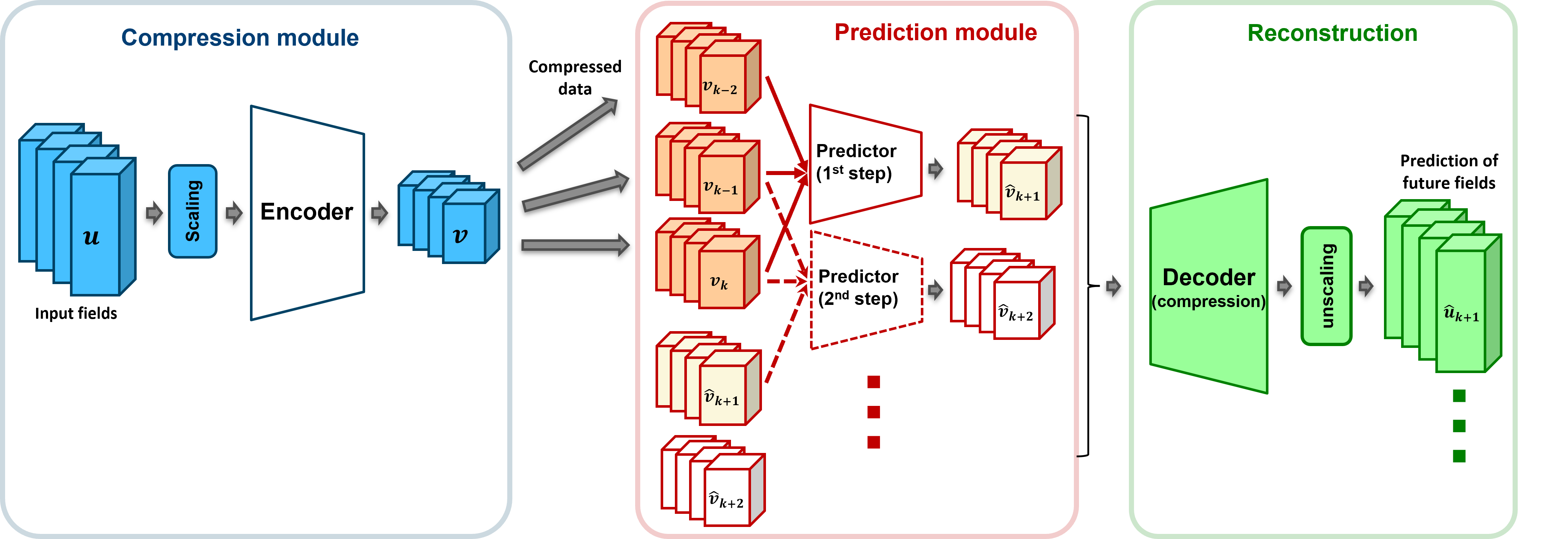}
\caption{Workflow for inference using one of the reduced-order surrogate models. The compression module produces an initial condition, the predictor uses the latent solution at three previous timesteps to generate a prediction for the next timestep, time marching is used to generate a sequence of predictions in latent space and, finally, the predictions are reconstructed to the original space.}\label{fig:SurrogCompressionWorkflow-Inference}
\end{figure}

When using this ROM framework to infer evolution of the concentration, porosity and velocities, initial conditions are encoded to latent space before passing them to the prediction network. Time marching is performed with the prediction network by predicting for a single timestep ($k+1$) from three previous timesteps ($k$, $k-1$ and $k-2$). This prediction ($k+1$) is then used with two previous timesteps ($k$ and $k-1$) to produce the next timestep ($k+2$) and so on. This generates a sequence of solutions that are uncompressed by the decoder and to which inverse scaling is applied to obtain the predictions in the physical or original space as seen in Figure~\ref{fig:SurrogCompressionWorkflow-Inference}. The time marching or autoregressive calling of the prediction network is represented in the figure by the dashed lines. 

\begin{figure}[!htb]
	\centering
	\includegraphics[width=\textwidth]{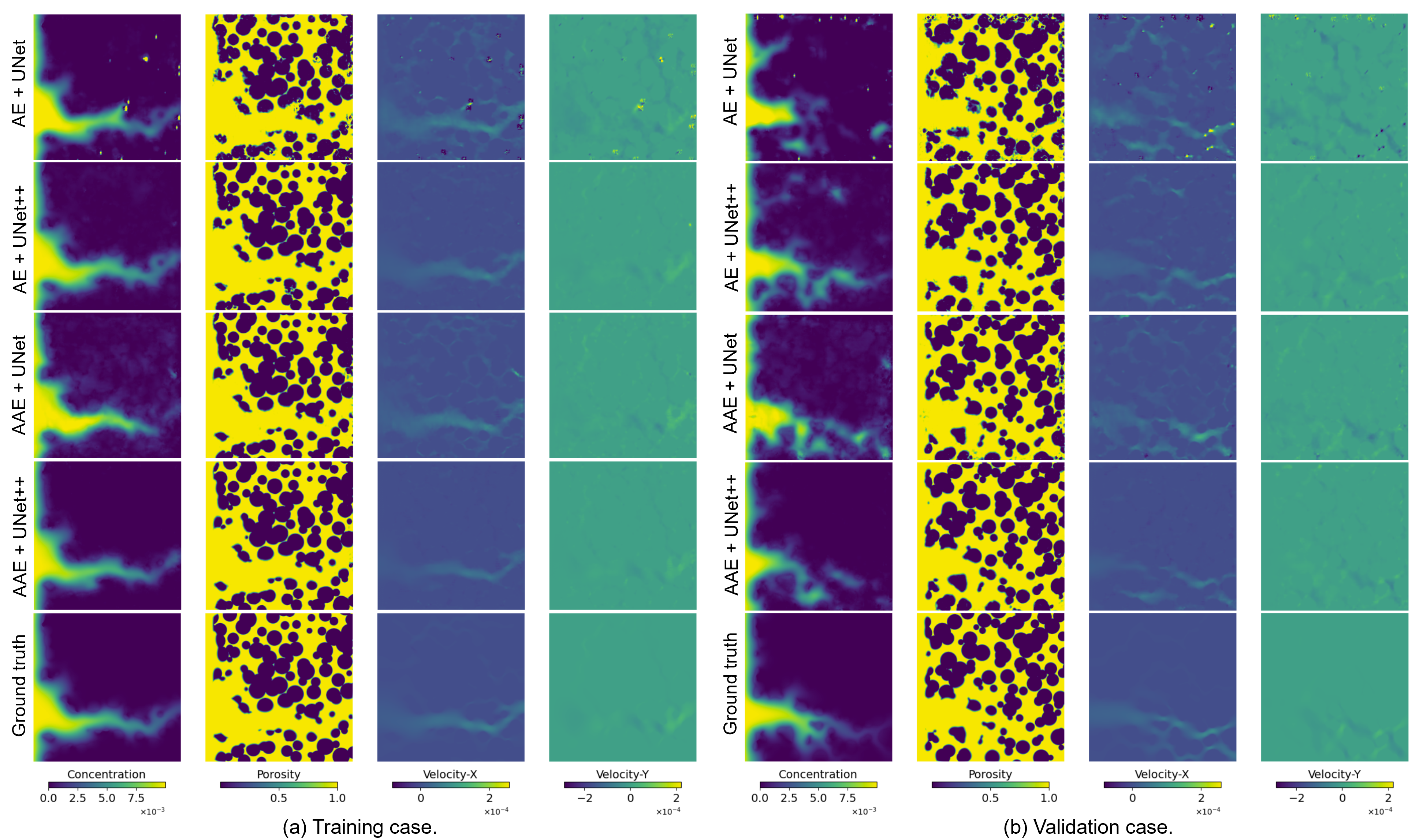}
	\caption{Surrogate model using compression - Autoregressive prediction after 100 timesteps in the carbon storage dataset, which contains four fields: concentration of CO2, porosity, and velocities in X- and Y-direction. Comparison of the results considering compression with an autoencoder or an adversarial autoencoder and prediction considering a UNet or a UNet++.}
	\label{fig:Results_Compression}
\end{figure}

Figure~\ref{fig:Results_Compression}(a) shows the results for predicting a sequence of 100 timesteps starting with one training sample taken from the carbon storage dataset. The figure compares the results with four different strategies for reduced-order surrogate models: for compression, we used a convolutional autoencoder, both with traditional training (AE) and with an adversarial training strategy (AAE). Using the two compressed datasets generated with these compressors, we tested a prediction network based on a UNet and on a UNet++. As illustrated, the predictions for the training samples have good accuracy for all cases analysed, with models using a UNet slightly worse than others. For the cases with UNet++, it is difficult to discern the differences between the predictions with training samples and the ground truth, that is, the result of the original GeoChemFOAM simulation. The models can predict the formation of the main channel and its bifurcations in the right direction and generally preserve the physical characteristics of the system after 100 timesteps (starting with an initial condition taken from the training data). 

By contrast, when observing the predictions starting with initial conditions taken from the validation dataset presented in Figure~\ref{fig:Results_Compression}(b), we can observe some divergence from the ground truth. In general, the predictions maintain reliability for some timesteps of rollout inference, but then start diverging from the ground truth, resulting in the images shown after 100 timesteps. Some degradation in porosity, far from the channels where the carbon dioxide is percolating, especially for the models with UNet, is also observed. For the first model (UNet without adversarial training in compression) we observe bright spots in the picture, indicating that the predictions in these regions are outside the ground-truth scale.

\begin{figure}[!htb]
	\centering
	\begin{subfigure}[t]{0.99\textwidth}
		\centering
		\includegraphics[width=\textwidth]{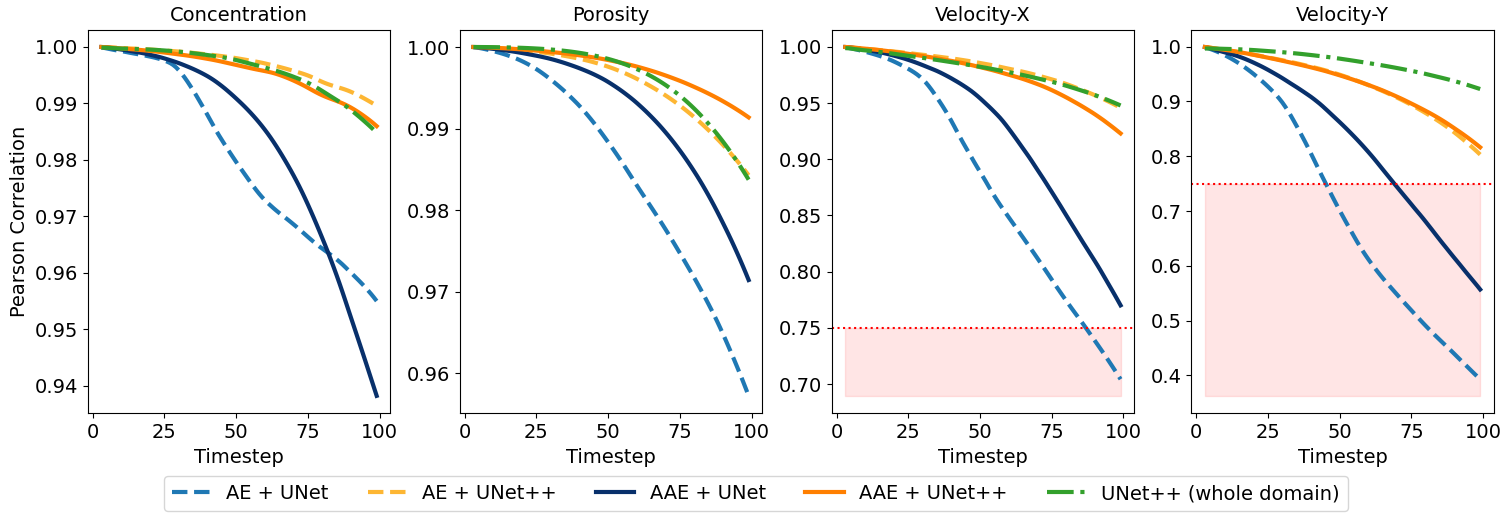}
		\caption{Average PCC per field for training simulations.}
		\label{fig:PCC_Compression_Train}
	\end{subfigure}  \;
	\begin{subfigure}[t]{0.99\textwidth}
		\centering
		\includegraphics[width=\textwidth]{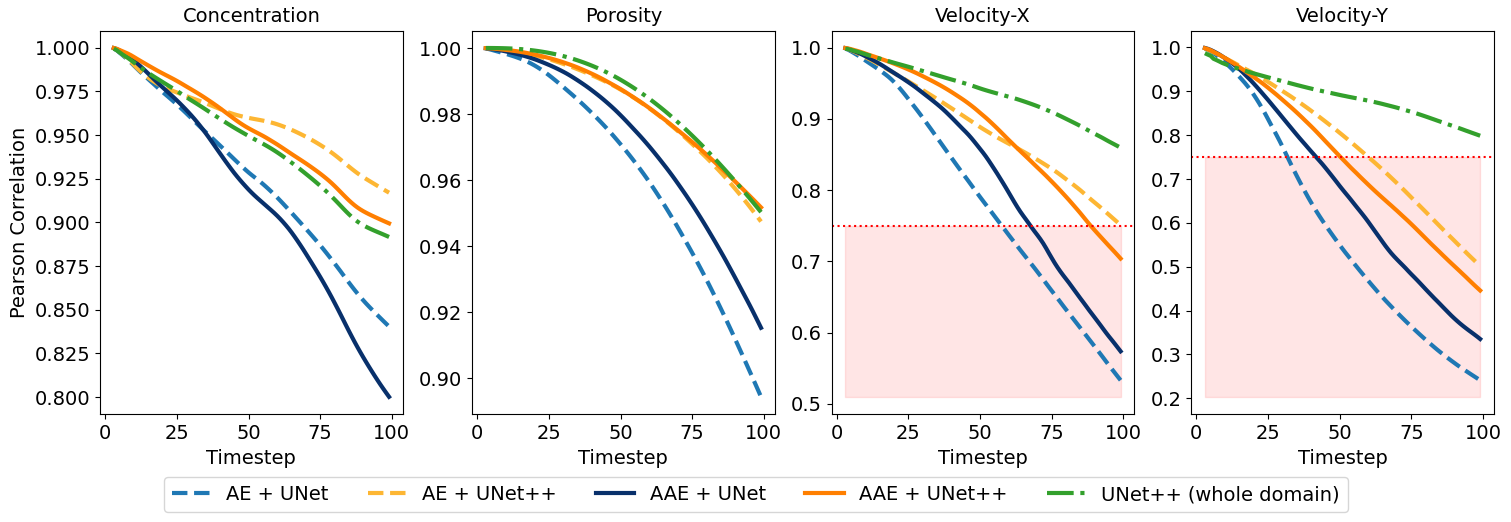}
		\caption{Average PCC per field for validation simulations.}
		\label{fig:PCC_Compression_Valid}
	\end{subfigure} \;
	
	\caption{Pearson correlation coefficient (PCC) for autoregressive prediction. Comparison of results obtained using compression with an autoencoder (dashed lines) or an adversarial autoencoder (solid lines) and prediction considering a UNet (blue lines) or a UNet++ (orange lines). The green line presents a base case of a UNet++ without ROM. For comparison, a shaded area is shown under $PCC=0.75$, indicating the timestep where the model is losing accuracy.}
	\label{fig:PCC_Compression}
\end{figure}

The evolution over time for each model can be better analysed in Figure~\ref{fig:PCC_Compression}, where each subplot shows the Pearson Correlation Coefficient for a field and the different lines correspond to each model, as shown in the legend. Each line shown in the figure is an average of the metrics obtained for all the simulations used for training (24 simulations, average shown in~\ref{fig:PCC_Compression_Train}) or validation (8 simulations, average in~\ref{fig:PCC_Compression_Valid}). As observed in Figure~\ref{fig:Results_Compression}, the models with a UNet predictor have the worst performance. The UNet++ models performed similarly, with excellent performance on the training dataset, but lower accuracy on the validation data after rolling out for some timesteps. With the UNet, the model with Adversarial Autoencoder outperformed the model with traditional training. With UNet++, though, the AAE outperformed only for the first timesteps, then its metrics start dropping faster, and, after roughly 50 timesteps, traditional training outperformed the AAE.

The figure also shows a comparison with a baseline model, which is a UNet++ trained on whole domain samples, that is, without any compression to reduce the data size. We observe that the reduced-order models with UNet++ perform similarly to the baseline for concentration of $\mathrm{CO_2}$ and porosity. However, the baseline outperforms the models with compression in the prediction of the velocities. As mentioned before, this work focuses on applications involving a large volume of data, where training and inference require substantial memory resources. The strategy of compressing or utilising the grid-size-invariant approach enables working with limited resources. As we will see in the next section, the grid-size-invariant approach performs similarly to this baseline whilst maintaining the improved computational performance.

Depending on the application desired, one model may be preferable to another. For instance, a common objective in modelling carbon dioxide flow in porous media is to predict rock dissolution. This is obtained by analysing the variation in the porosity field relative to the original field immediately before injection starts. In this case, it is recommended to use any of the models with UNet++, as they performed similarly and outperformed the others for the porosity field. In contrast, suppose the goal is to combine the surrogate model with a PDE solver to accelerate simulations. In this case, it is recommended to use surrogate models to predict a smaller window of timesteps, thereby keeping the metrics above a threshold and avoiding degradation of the velocity fields. For instance, if we choose a threshold of $PCC=0.75$, the Pearson correlation remains above this value for up to 30 timesteps with AE+UNet, whereas AAE models remain above the threshold for roughly 50-60 timesteps.

\subsection{Grid-size-invariant surrogate models}\label{subsecR-Grid-inv}

\begin{figure}[!htb]
	\centering
	\includegraphics[width=\textwidth]{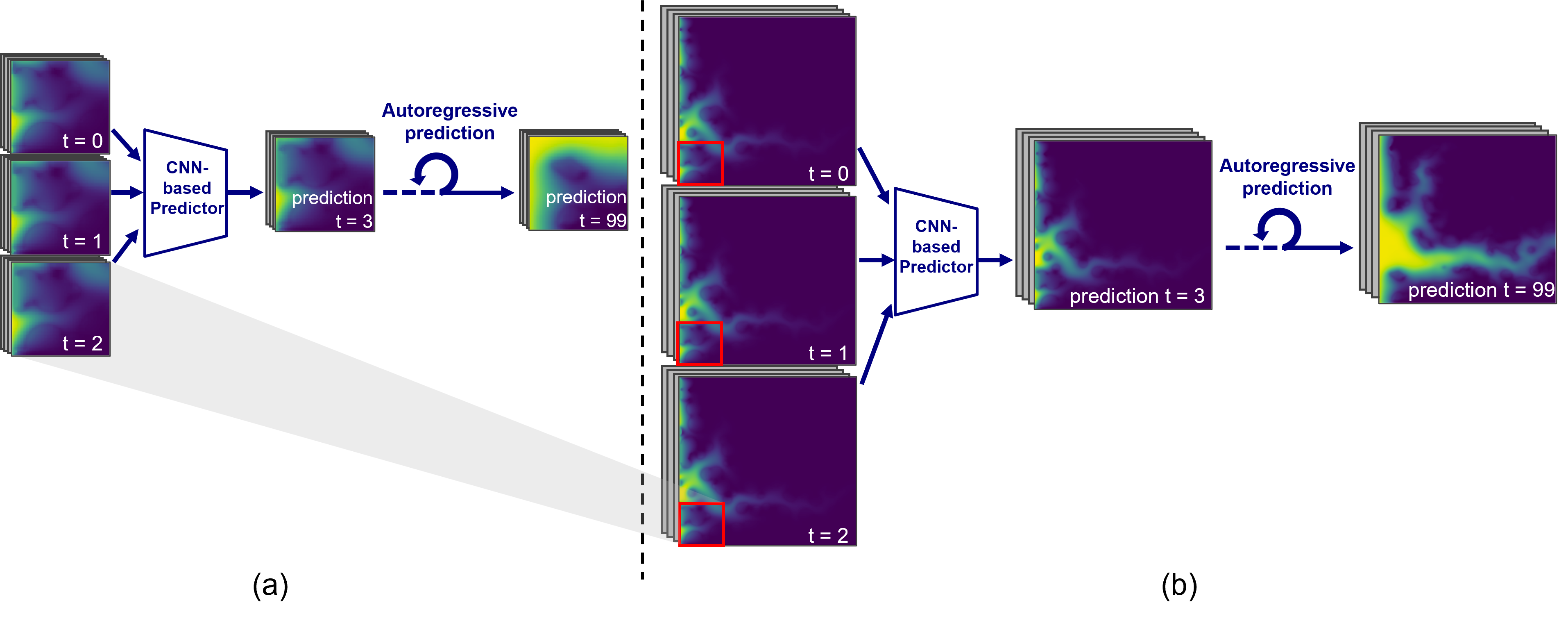}
	\caption{Schematic for the grid-size-invariant framework. (a) Prediction with a $64 \times 64$ sample used for training. After training, the predictor can be used with these samples to generate an autoregressive prediction for a 100-timestep sequence. (b) In the grid-size framework, the same predictor can be used to do a prediction with a $256 \times 256$ image (here showing images from the training dataset; red squares represent the $64 \times 64$ samples shown in (a)).
    }
	\label{fig:Diagram_with_Results_GI}
\end{figure}

To improve results whilst maintaining adaptability to large-domain datasets, we implemented the Grid-Size-Invariant framework proposed here for the carbon storage scenario, using a single network for prediction without prior compression. As mentioned in Section~\ref{subsecMet-GridInv}, a CNN-based network can capture spatial features and a fully convolutional network can be applied to domains of any size. Consequently, the training may be performed on smaller representative domains, and the network can predict over larger unseen domains. Here, we used $64 \times 64$ samples for training and inferred over the whole spatial domain of the dataset, i.e., $256 \times 256$. The samples were uniformly distributed in the spatial domain and sampled across all timesteps available for each simulation in the training data (similar to the ROM surrogates, 24 simulations from the dataset were used for training, whilst 8 were reserved for validation). Figure~\ref{fig:Diagram_with_Results_GI} shows a schematic of a grid-size framework considering samples of the training dataset. Figure~\ref{fig:Diagram_with_Results_GI}(a) contains inputs of $64 \times 64$ used during training. For illustrative purposes, we show only the concentration field here, but the model input also considers the other three fields over the three consecutive timesteps, as represented by the grey squares stacked below the concentration field. The output is a prediction for the next timestep of the four fields. The figure also shows that the trained neural network can later be used for autoregressive prediction, as shown in the prediction module of Figure~\ref{fig:SurrogCompressionWorkflow-Inference}, in order to generate a forecast for a longer sequence of timesteps. Furthermore, the grid-size invariance allows us to use the predictor with larger domains, as shown in Figure~\ref{fig:Diagram_with_Results_GI}(b), where we use the trained predictor to evolve the sequence in the whole images of $256 \times 256$. The red square shows the subsample used in (a).

\begin{figure}[!htb]
	\centering
	\includegraphics[width=\textwidth]{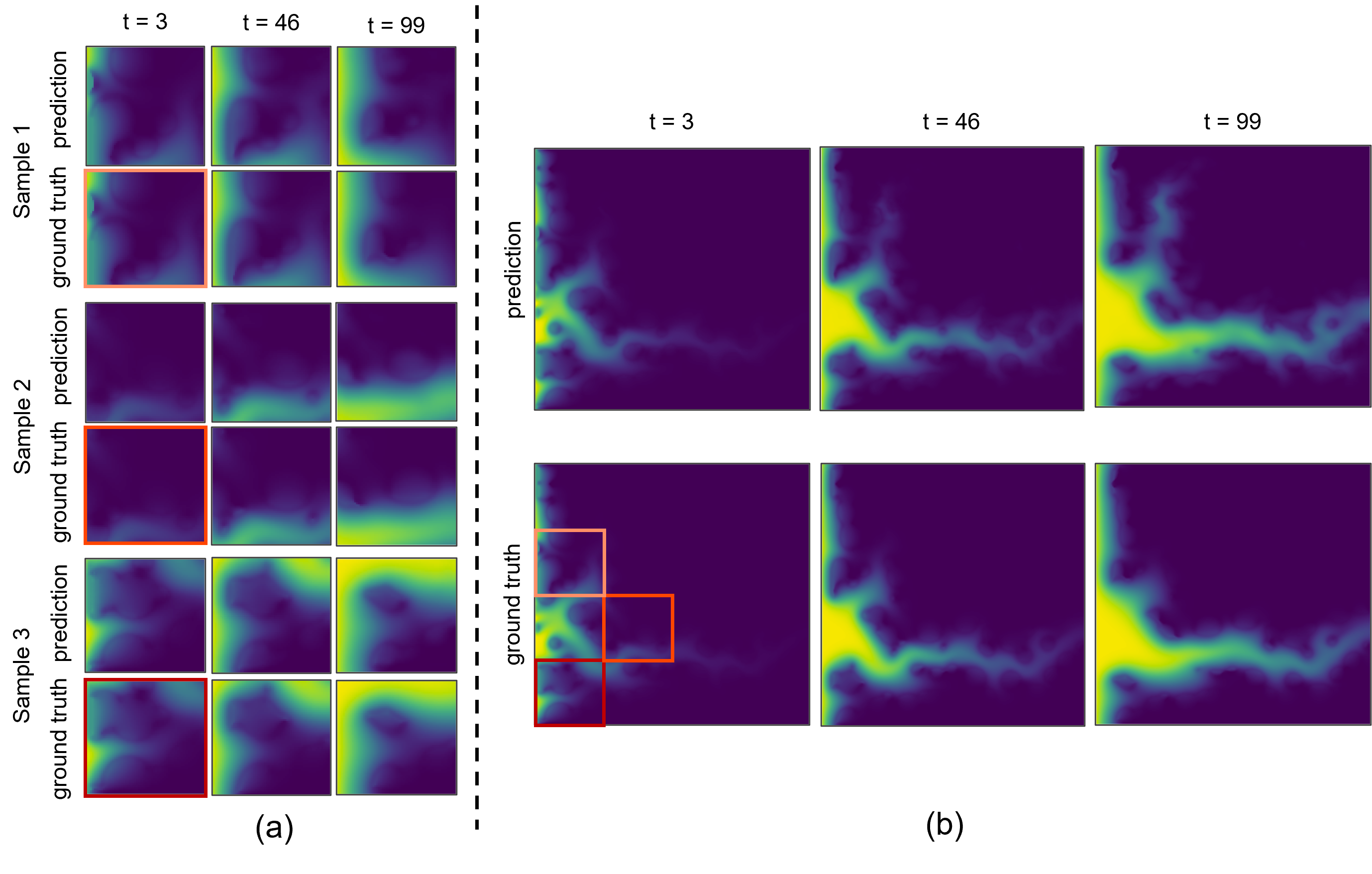}
	\caption{(a) Prediction with a $64 \times 64$ sample used for training. (b) In the grid-size framework, the same predictor can be used to do a prediction with a $256 \times 256$ image (images from training dataset; red squares represents the $64 \times 64$ samples shown in (a)). }
	\label{fig:GSI_Results_TrainingData}
\end{figure}

The results for three different samples of the same original simulation are shown in Figure~\ref{fig:GSI_Results_TrainingData}. In Figure~\ref{fig:GSI_Results_TrainingData}(a) we can observe the results with three samples used in training. We show the predictions of timestep $t=3$, which correspond to a single forward pass through the trained predictor, and the results of autoregressive prediction for the 46th timestep and the 99th timestep. Similarly, Figure~\ref{fig:GSI_Results_TrainingData}(b) shows the result of a single forward pass through the trained predictor ($t=3$) and the autoregressive prediction for $t=46$ and $t=99$, although now the inputs for the predictor were the original domain, that is, $256 \times 256$ images. The good performance obtained in (b) is due to the grid-size-invariant property of the framework used. 

As in the case of the predictor for ROMs, for our single grid-size-invariant neural network, we compared the results obtained with UNet and UNet++ architectures, using the same layers presented in Figure~\ref{fig:Pred_UNetvsUNetpp}. Additionally, we compare the results of traditional training with the rollout training strategy, as explained in Section~\ref{subsecMet-Rollout}, considering $T=8$ timesteps rolled out in the training loop. For the results shown in Figure~\ref{fig:GSI_Results_TrainingData}, we used the trained model with UNet and rollout. 

\begin{figure}[!htb]
	\centering
	\includegraphics[width=\textwidth]{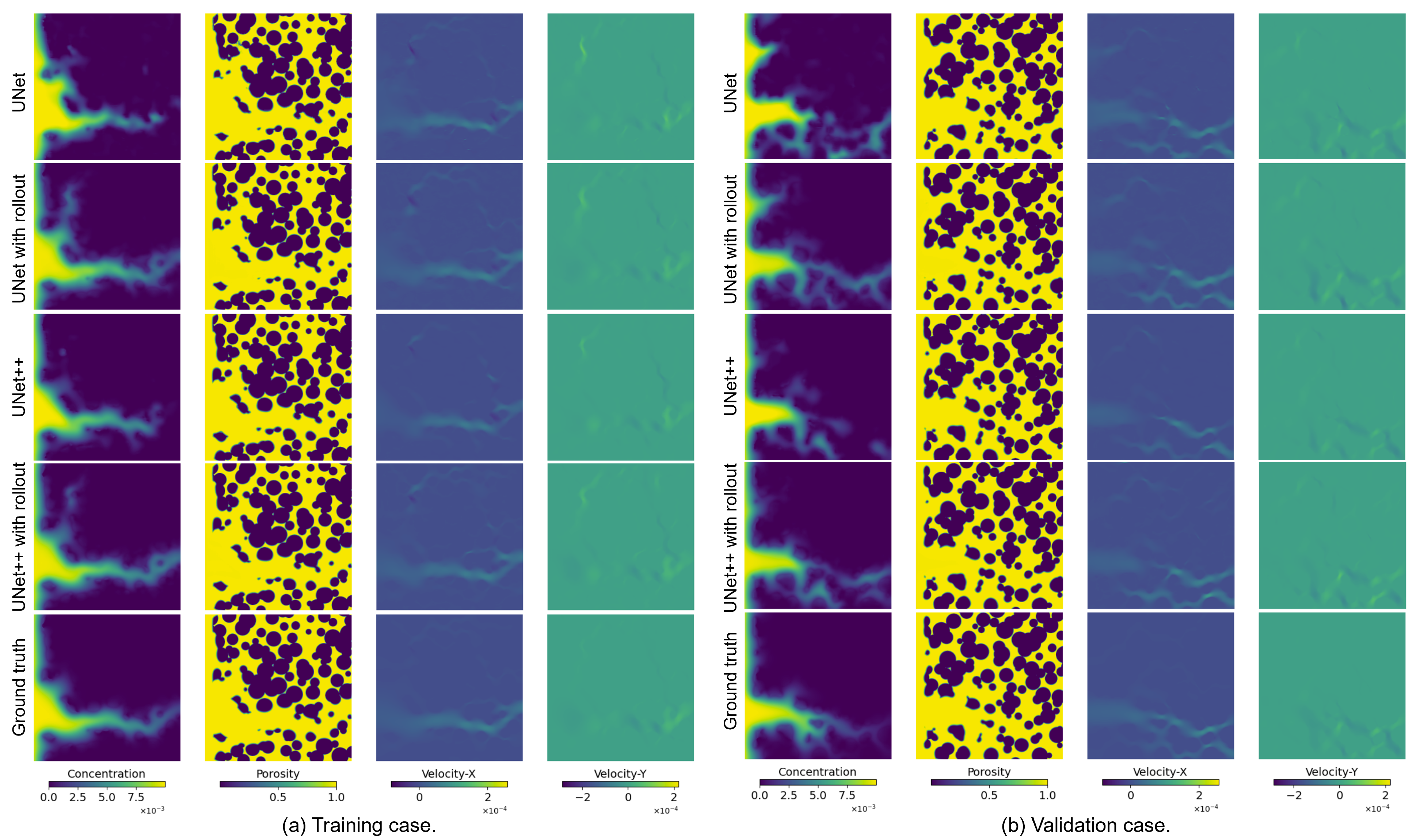}
	\caption{Surrogate model using grid-size invariance (results obtained for a $256 \times 256$ domain, although the network was trained with $64 \times 64$ samples) - comparison of autoregressive prediction after 100 timesteps considering a UNet or a UNet++, both with traditional training and with rollout training.}
	\label{fig:Results_GI}
\end{figure}

The comparison of the autoregressive prediction of 100 timesteps in the four Grid-Size-Invariant surrogates is in Figure~\ref{fig:Results_GI}, where (a) presents the results for one case with a $256 \times 256$ domain, for which subsamples were used for training the network, and (b) shows the result for one validation case, which is an unseen case (also a $256 \times 256$ domain). As shown in the picture, the predictions yield reasonable results for all models, accurately detecting the formation of the main channel and maintaining the porosity field without any changes where the flow has not yet reached. Notably, UNet++ outperforms UNet models. We also observe improvements in both architectures when using rollout training. 
For the validation simulation presented, the surrogates based on UNet predict a secondary channel forming upward after the autoregressive prediction of many timesteps, whereas the ground truth does not contain this channel. This effect was reduced with rollout training and disappeared with UNet++, showing that the complexity added in the architecture improved the results. Overall, the models with rollout outperformed the corresponding ones without rollout, and the UNet++ outperformed the UNet.

\begin{figure}[!htb]
	\centering
	\begin{subfigure}[t]{0.99\textwidth}
		\centering
		\includegraphics[width=\textwidth]{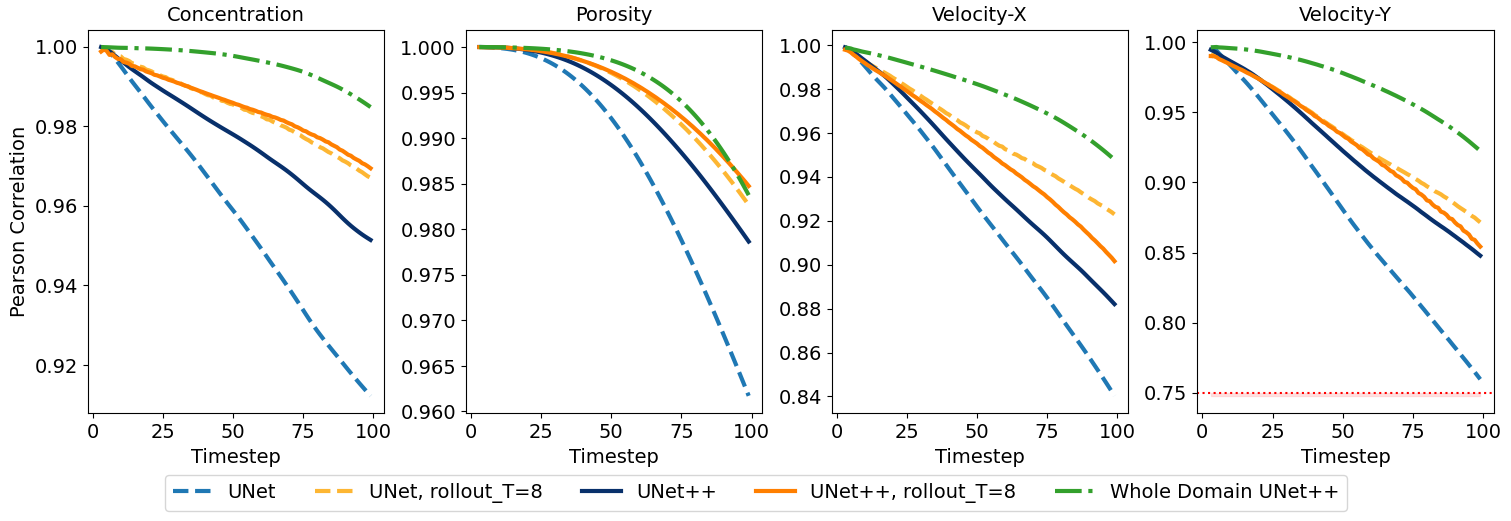}
		\caption{Average PCC per field for training simulations.}
		\label{fig:PCC_GI_Train}
	\end{subfigure}  \;
    
	\begin{subfigure}[t]{0.99\textwidth}
		\centering
		\includegraphics[width=\textwidth]{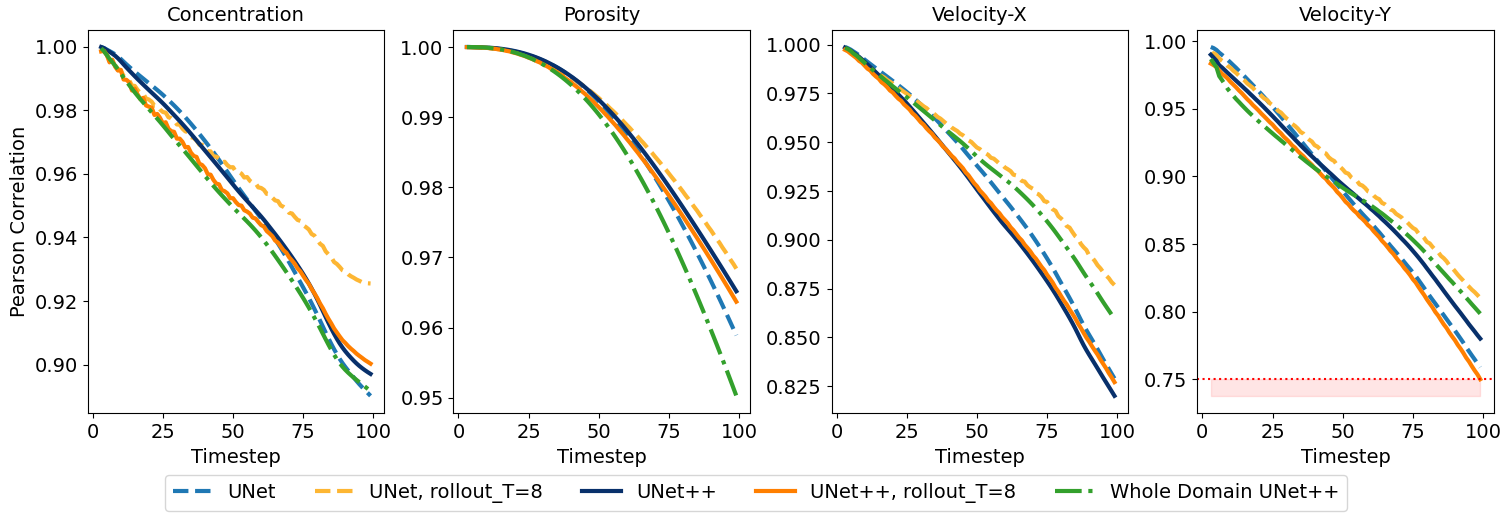}
		\caption{Average PCC per field for validation simulation.}
		\label{fig:PCC_GI_Valid}
	\end{subfigure} \;
	
	\caption{Pearson correlation coefficient (PCC) for autoregressive prediction. Comparison of results obtained using the grid-size-invariant framework with a UNet or a UNet++ and considering traditional training and rollout training with $T=8$ timesteps rolled during training.}
	\label{fig:PCC_GI}
\end{figure}

To analyse the general results with all the simulations available, we calculated the temporal evolution for the Pearson Correlation Coefficient (PCC) for each simulation (24 simulations were used for training and 8 remained for validation).
Figure~\ref{fig:PCC_GI} illustrates the temporal evolution of the average PCC per field of each grid-size-invariant model, as shown in the legend. For comparison, a UNet++ trained on the entire domain samples, used as a baseline model here, is also presented. It is worth noting that the model trained on the whole domain and using a UNet network performed slightly worse than the one with UNet++. This demonstrates that UNet++, as a more complex network, requires more samples to achieve significant performance gains over UNet.

The comparison with the baseline model shows that the grid-size-invariant frameworks do not perform as well as the baseline for the simulations used during training. However, they outperform the baseline on unseen data, demonstrating that the grid-size-invariant method leverages data augmentation enabled by domain subsampling and reduces model overfitting. This data augmentation also helped UNet++ to achieve a better performance, resulting in a significant advantage over UNet in the grid-size-invariant framework.

Among the Grid-Size-Invariant models, the models performed similarly, but the models with rollout training slightly outperformed the ones without it. The Pearson correlation remains above a threshold of 0.75 for the 100-timestep prediction in all fields. 

Additionally, when observing images from the inference, our perception sometimes diverge from the results expected when comparing PCC values. Particularly, for the models with UNet++ the rollout does not seem to bring an improvement in PCC metrics, but we can see some improvements in the images. Therefore, we present in Table~\ref{tab:results_GI_pcc_ssim} the results of the Structural Similarity Index Measure (SSIM), which captures the human perception when comparing images. The values of these metrics here represent the average value for all the validation simulations in the last timestep. The SSIM results highlight the advantage of combining UNet++ with the rollout training strategy to enhance the model, in accordance with the perception when observing the resulting images of the fields. Using both metrics, the best model overall was UNet with rollout, closely followed by the others.

\setlength{\tabcolsep}{3.4pt}  % default is usually 6pt
\begin{table}[h]
\centering
\caption{Pearson Correlation Coefficient (PCC) and Similarity (SSIM) metrics per field for different models using the Grid-Size-Invariant framework. Green values indicate the best performances.}
\renewcommand{\arraystretch}{1.2}
\begin{tabular}{l|cccc|cccc}
\toprule
 & \multicolumn{4}{c|}{\textbf{PCC}} & \multicolumn{4}{c}{\textbf{SSIM}} \\ 
\midrule
\textbf{Model} & $C_{\mathrm{CO_2}}$ & Porosity & Velocity-X & Velocity-Y & $C_{\mathrm{CO_2}}$ & Porosity & Velocity-X & Velocity-Y \\ 
\midrule
UNet & 0.89 & 0.96 & 0.83 & 0.76 & 0.79 & 0.89 & 0.93 & 0.94 \\
UNet rollT8 & \textcolor{green!60!black}{\textbf{0.93}} & \textcolor{green!60!black}{\textbf{0.97}} & \textcolor{green!60!black}{\textbf{0.88}} & \textcolor{green!60!black}{\textbf{0.81}} & 
\textcolor{green!60!black}{\textbf{0.81}} & 
\textcolor{green!60!black}{\textbf{0.91}} & 
\textcolor{green!60!black}{\textbf{0.94}} &  
\textcolor{green!60!black}{\textbf{0.95}} \\
UNet++ & 0.90 & \textcolor{green!60!black}{\textbf{0.97}} &
0.82 & 0.78 & 0.80 & 
\textcolor{green!60!black}{\textbf{0.91}} & 0.92 & 0.94 \\
UNet++ rollT8 & 0.90 & 0.96 & 0.83 & 0.75 & 
\textcolor{green!60!black}{\textbf{0.81}} & 0.90 & 0.93 & \textcolor{green!60!black}{\textbf{0.95}} \\
\bottomrule
\end{tabular}
\label{tab:results_GI_pcc_ssim}
\end{table}
\setlength{\tabcolsep}{6pt}  % default is usually 6pt

Table~\ref{tab:results_GI_MSE_areaCO2} complements the metrics presented before. It shows the Mean Squared Error (MSE) obtained when comparing the scaled predicted fields after 100 timesteps and the ground truth values for the validation cases. It also shows the error in the prediction of the area occupied by $\mathrm{CO_2}$, calculated following the equation~\ref{eq_areaCO2}. This metric was calculated in the concentration field scaled using min-max scaling, and we chose $C_\text{threshold}=0.5$. The values presented are the median, first quartile Q1 and third quartile Q3 for the cases used for validation. This metric also highlights the advantage of using rollout training to reduce prediction errors that accumulate over multiple timesteps, and its low absolute values indicate that the proposed models maintain good statistics and are robust for multi-query problems, such as uncertainty quantification. Again, these metrics show that the UNet with rollout slightly outperformed other models.

\begin{table}[h]
\centering
\caption{MSE and error in area occupied by $\mathrm{CO_2}$ for different models using the Grid-Size-Invariant framework.}
\renewcommand{\arraystretch}{1.2}
\begin{tabular}{lcccc}
\toprule
 &  & \multicolumn{3}{c}{\textbf{Error Area $\mathrm{CO_2}$}}\\ 
%\midrule
\cmidrule{3-5}
\textbf{Model} & {\textbf{MSE}}  & Q1 & Median & Q3 \\ 
\midrule
UNet & 0.0094 & -2.3 & -1.8 & -1.3 \\
UNet rollT8 &  \textcolor{green!60!black}{\textbf{0.0063}} & 0.1 & \textcolor{green!60!black}{\textbf{1.1}} & 1.3 \\
UNet++ & 0.0078 & 1.8 & 2.1 & 2.8 \\
UNet++ rollT8 & 0.0076 & 0.7 & 1.6 & 2.5 \\
\bottomrule
\end{tabular}
\label{tab:results_GI_MSE_areaCO2}
\end{table}

\begin{figure}[!htb]
	\centering
	\includegraphics[width=\textwidth]{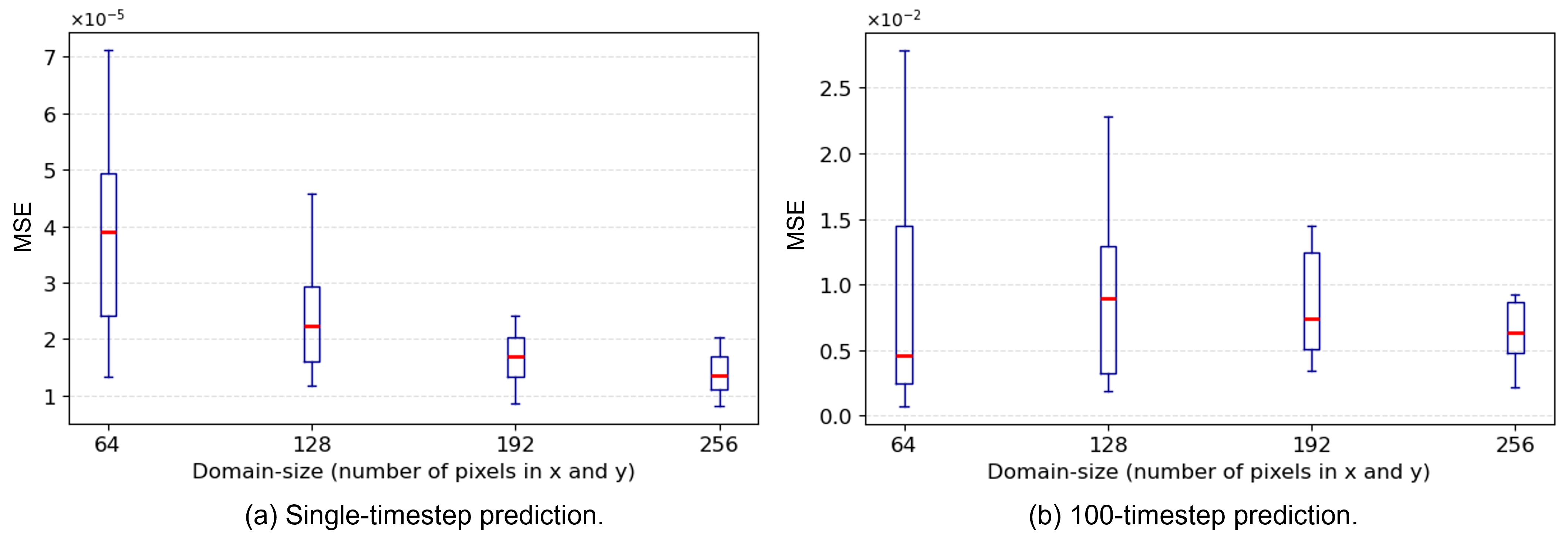}
	\caption{Grid-size invariance: these plots show the distribution of mean squared error for inference using different domain sizes for (a) a single-timestep prediction; and (b) the autoregressive prediction for the 100th timestep.}
	\label{fig:MSE_different_sizes}
\end{figure}

Figure~\ref{fig:MSE_different_sizes} demonstrates the grid-size property, that is, the capacity to generalise the inference over bigger domains. The box plot shows the distribution of errors when inference is performed with different domain sizes using validation data. Figure~\ref{fig:MSE_different_sizes}(a) shows the results for a single-timestep prediction, whereas (b) shows how it evolves when using the autoregressive inference for predicting a sequence of 100-timesteps. This figure was made considering samples on the left side of the images, which contain $x0=0$, where the flow starts. Thus, for $64 \times 64$ we have 4 samples uniformly distributed in y; $128 \times 128$ contains 2 samples per image, uniformly distributed in y; $192 \times 192$ also contains 2 samples per image, but with some overlap; $256 \times 256$ is only one sample per image. Figure~\ref{fig:MSE_different_sizes}(a) shows a decreasing error when increasing the size of the prediction. This happens because larger domains contain more pixels that are not affected by the flow, making prediction easier. When considering the autoregressive long-term prediction, this effect disappears, as shown in Figure~\ref{fig:MSE_different_sizes}(b), which highlights how the median behaviour of the error remains in the same range for different sizes of input.

\subsection{General comparison of methods}\label{subsecR-Comparison}

The Table~\ref{tab:Memory_Time_Comparison} presents a comparison of the frameworks presented here in terms of memory consumption for training and time to train and infer. For each framework, the table depicts the total number of parameters in the prediction network, indicating the complexity of the network. Therefore, it is directly correlated with the time for training, also shown in the table. The time for training shown there considers the number of epochs necessary for a good convergence of training/testing losses. In the case of rollout training models, as we initialised the training with the weights obtained without rollout, this means only a few epochs. All the simulations were run using one GPU NVIDIA GeForce RTX 4060.

Concerning the ``Time for inference'' column, it refers to the time for autoregressive prediction of 97 timesteps, that is, using the first three timesteps of a simulation and predicting the total timesteps we have in our dataset.

Additionally, it is worth highlighting that in ``Time for training'', we show the time for the models of each framework for which we showed the results in this paper. However, the process of hyperparameter tuning consumes a significant amount of time running other models. In this sense, adversarial training represents a disadvantage when compared to the hyperparameter tuning in a traditional autoencoder, as we can change three different learning rates to maintain the fooling game between the encoder and the discriminator. We also tested different ratios of updates in these learning rates, as described in the Appendix~\ref{ap:compression}.

\begin{table*}[h]
\centering
\caption{Comparison of methods in memory and time consumption. Time for training in the ROM models presents the time for the compression network (C) and for the prediction network (P). Time for inference refers to the time for autoregressive prediction of 97 timesteps.}
%\resizebox{\textwidth}{!}{%
%\begin{tabular}{@{}L{1.3cm}L{2.3cm}C{2cm}C{2.2cm}C{2.5cm}C{1.8cm}@{}}
\begin{tabular*}{\textwidth}{@{\extracolsep\fill}llcccc}
%\begin{tabular*}{\textwidth}{@{\extracolsep{-5pt}}llcccc}
\toprule

\makecell[c]{} & 
\makecell[c]{\textbf{Method}} & 
\makecell[c]{\textbf{Total} \\ \textbf{Parameters}} &  
\makecell[c]{\textbf{GPU-peak} \\ \textbf{memory for} \\ \textbf{training [GB]}} & 
\makecell[c]{\textbf{Time for} \\ \textbf{training} \\ \textbf{[min]}} & 
\makecell[c]{\textbf{Time for} \\ \textbf{inference} \\ \textbf{[sec]}} \\ 
\midrule
\multirow{2}{*}{\makecell[l]{Whole \\Domain}} 
 & UNet & 7.7M & 1.809 & 191 & $<1$ \\
 & UNet++ & 9M  & 3.444 & 669 & $\sim 1.5$ \\ 
\midrule
\multirow{4}{*}{ROMs}%{\makecell[l]{Compression \\+ Prediction}} 
 & AE + UNet & 855k + 7.7M & 0.263 & 399(C)+80(P) & $<1$ \\
 & AE + UNet++ & 855k + 9M  & 0.395 & 399(C)+266(P) & $\sim 1.5$ \\
 & AAE + UNet & 856k + 7.7M & 0.263 & 534(C)+144(P) & $<1$ \\
 & AAE + UNet++ & 856k + 9M & 0.395 & 534(C)+210(P) & $\sim 1.5$ \\ 
\midrule
\multirow{4}{*}{\makecell[l]{Grid-Size \\Invariant}} 
 & UNet & 7.7M & 0.263 & 194 & $<1$ \\
 & UNet rollT8 & 7.7M & 0.897 & 183 & $<1$ \\
 & UNet++ & 9M   & 0.393 & 222 & $\sim 1.5$ \\
 & UNet++ rollT8 & 9M  & 1.542 & 142 & $\sim 1.5$ \\ 
\bottomrule
\end{tabular*}%
%}
%\footnotetext[1]{Run in 1 GPU NVIDIA GeForce RTX 4060.}
\label{tab:Memory_Time_Comparison}
\end{table*}

The comparison of GPU peak memory shows the advantage of training the prediction model in a reduced-order space or using smaller subdomains based on the grid-size-invariant approach. This enables training models with large datasets using computational resources that can not afford training models with the whole domain. Additionally, we can observe that the UNet++ requires approximately $1.5\times$ to $2 \times$ the memory needed for the models with UNet. 

Finally, inference in these data-driven models is extremely fast and represents a gain of several orders of magnitude compared with a PDE-based approach. The time to produce each simulation in GeoCHemFOAM used to generate this dataset was approximately 3 hours using 24 CPUs of 3 GHz each~\cite{Cirne2025_DL_IterativeStacked_CO2}.

\section{Conclusions}\label{secConclusions}

This work compares different strategies for developing a machine-learning-based surrogate model for rock-fluid interaction prediction. The first group of frameworks contains reduced-order models based on autoencoders for compression and a UNet or UNet++ for prediction. Then, we propose a grid-size-invariant approach and compare several models with this approach. The main findings are listed below.

\begin{enumerate}
    \item Adversarial training for compression highly increases the complexity for training, but it brought a gain in comparison with the same autoencoder without adversarial training. Even if the generator stops fooling the discriminator after some epochs, the use of the adversarial training strategy to start the training helps to find a more regularised latent space for compression, and this improves the predictions in the latent space.

    \item When enough data to train a more complex network is available, UNet++ resulted in better predictions than UNet. Although the UNet++ was first proposed for image segmentation, this work shows its strength for surrogate modelling.

    \item The grid-size-invariant framework outperformed the Reduced Order Model approach for unseen data, as observed in Pearson Correlation Coefficient comparisons and also in fields visualisation of the evolution of autoregressive prediction.

    \item Compression reduces memory and time consumption during predictor training and inference. Although it is necessary to train the compression network first and it requires the entire field's images, sampling in time can be used. The compression network should require less data than the predictor, as it only needs to learn spatial features, not the time evolution. In contrast, the grid-size-invariant network only reduces memory during training, and uses the whole field's images during inference.

    \item Rollout training helped to obtain surrogate models with better performance for long autoregressive prediction.
\end{enumerate}

One of the key considerations when selecting methods for comparison was the feasibility of applying them to large-domain datasets, particularly 3D datasets. Thus, the frameworks compared here considered reductions in memory during training to allow the use with large datasets. As future work, we will apply the proposed methods to 3D datasets.

Also, in the future, we will test the grid-size-invariant approach in a variety of fluid flow problems governed by different combinations of PDEs. As a data-driven method, we aim to demonstrate its applicability to potentially any fluid flow simulation, provided that we choose a subsample large enough to capture the flow patterns.

Although this work focuses on carbon injection, the algorithms developed here extend beyond the presented example, offering broader applicability to image-based modelling of time-evolving systems.

Finally, the surrogate will be combined with a PDE-based solver (potentially GeoChemFOAM, used to generate the original dataset). The surrogate will be used to predict for a number of timesteps in the simulation, whilst reverting to the PDE-solver if the metrics fall below a certain threshold. Thus, we can benefit from the gains in efficiency provided by the surrogate and use the PDE-solver to guarantee the reliability of the prediction over a long-term run.

%%%%%%%%%%%%%%%%%%%%%%%%%

%\backmatter

\section*{Acknowledgments}
%\bmhead{Acknowledgements}

The authors would like to thank Petróleo Brasileiro S.A. (Petrobras) for sponsoring the doctoral research of Nathalie Carvalho Pinheiro. 
The authors would like to acknowledge the following EPSRC and NERC grants: ECO-AI, ``Enabling $\mathrm{CO_2}$ capture and storage using AI'' (EP/Y005732/1); the PREMIERE programme grant, AI-Respire, ``AI for personalised respiratory health and pollution (EP/Y018680/1); ``AI to enhance manufacturing, energy, and healthcare'' (EP/T000414/1); 
SMARTRES, ``Smart assessment, management and optimisation of urban geothermal resources'' (NE/X005607/1); AI4Urban-Health, ``AI Solutions to Urban Health Using a Place-Based Approach'' (APP55547 / UKRI 1241); and WavE-Suite, ``New Generation Modelling Suite for the Survivability of Wave Energy Convertors in Marine Environments'' (EP/V040235/1).  
Support from Imperial-X's Eric and Wendy Schmidt Centre for AI in Science (a Schmidt Futures program) is gratefully acknowledged.

\begin{appendices}

\section{Compression tuning and results}\label{ap:compression}

The network architectures presented in this work exhibit the same encoder and decoder layers for both Autoencoder and Adversarial Autoencoder; the unique difference lies in the presence of the discriminator, also known as the critic, which is an auxiliary network used only during training. The result of adversarial training is finalising the training with different weights for the encoder and decoder, and this should guarantee that the distribution in the latent space is adherent to the distribution used in training, in this case, a Gaussian distribution.

After testing different network architectures and extensive hyperparameter tuning, the encoder developed in this research comprises five layers. Two of these consider a stride of 2 with a kernel size of 2 by 2, which reduces the size while also changing the number of channels. The remaining three layers, with a stride of 1 and a kernel size of 3 by 3, are interspersed among the other two, changing the number of channels in the layer without decreasing the dimension of the input. So, these layers can help to smooth the changes in the number of channels. For the AAE, the discriminator comprises two convolutional layers. The total number of parameters is $854,728$ for AE and $856,201$ for AAE. The algorithms used for compression result in a reduction by a factor of 4 in each dimension of the 2D dataset, leading to a memory reduction with a ratio of $16:1$. 

For the Autoencoder, the main tuning focused on comparing the best network architectures, and the main hyperparameters used in the results presented here are shown in Table~\ref{tab:AEconfig}. We also tested different sampling and data augmentation by flipping the images upside-down. 

\begin{table}[h!]
\caption{Parameters for Compression AE.}
\begin{center}
\begin{tabular}{@{}cc@{}} %{*{2}{c}}
\toprule
Hyperparameter & Value \\
\midrule
Learning Rate (LR) & 0.001 \\
Optimizer & Adam \\
Adam parameter $\beta_1$ & 0.9 \\
Adam parameter $\beta_2$ & 0.999 \\
\bottomrule
\end{tabular}\label{tab:AEconfig}
\end{center}
\end{table}

Although the Adversarial Autoencoder has the advantage of regularising the latent space, it is much more challenging to tune its hyperparameters. We started with the same network approved for the Autoencoder and adjusted the learning rates of the Autoencoder, the Discriminator, and the Encoder in adversarial mode to maintain the adversarial training for as long as possible. However, we did not afford to maintain the adversarial game and reduce the loss for the Autoencoder to similar values to those obtained without adversarial training. Thus, the results presented here consider the adversarial training, but after $140$ epochs, the encoder loss rocketed and the discriminator loss tended to zero, indicating that the encoder was unable to continue fooling the discriminator. Nonetheless, we continued the training, and the autoencoder loss decreased faster after that, achieving similar values to those obtained without adversarial training. We stopped the training when the loss of testing samples was no longer significantly reducing. Maintaining adversarial training for more than $100$ epochs was only possible when considering some recommendations from previous works with AAEs or GANs, such as the use of Adam Optimiser with $\beta_1 = 0.5$, that is, reduce the exponential decay for the first momentum term (mean of the gradients) in Adam Optimizer when compared to its default value. This recommendation was highlighted in~\cite{radford2016_GAN}, where the authors comment that the default value of $0.9$ has led to training oscillation and instability, while reducing it to $0.5$ helped stabilise training. Additionally, we considered a lower update for the encoder in relation to the discriminator, as recommended in~\cite{GANsinAction_Langr2019}. The results showed here considered a ratio of $2:1$ between the discriminator and encoder updates. The Table~\ref{tab:AAEconfig} summarises the AAE parameters.

\begin{table}[h!]
\caption{Parameters for Compression AAE.}
\begin{center}
\begin{tabular}{@{}cc@{}} %{*{2}{c}}
\toprule %\hline
Hyperparameter & Value \\
\midrule %\hline
LR Autoencoder & 0.0005 \\ 
LR Discriminator & 0.00025 \\
LR Encoder (Generator) & 0.0005 \\
Discriminator/Encoder update ratio & $2:1$ \\
Optimizer & Adam \\
Adam parameter $\beta_1$ & 0.5 \\
Adam parameter $\beta_2$ & 0.999 \\
\bottomrule %\hline
\end{tabular}\label{tab:AAEconfig}
\end{center}
\end{table}

Here, we present the results of the reconstruction using the compressor autoencoder (AE) and the adversarial autoencoder (AAE), and compare them with two other downsampling methods commonly used in image processing with the same reduction ratio. The first one is the function cv.Resize, available in the OpenCV library. It offers various algorithm options. Here, we used downsampling with an interpolation based on the area relation and upsampling based on a bicubic interpolation over a 4x4 pixel neighbourhood. The second image reduction method implemented for comparison is the use of Gaussian Pyramid downsampling and upsampling. They were implemented using the functions cv.pyrDown and cv.pyrUp. The concept behind this algorithm is to implement a convolution with a Gaussian kernel to obtain a downsampled image. However, when applying the Gaussian kernel to upsampling, it creates a blur effect in the reconstructed image. Another upsampling method could generate better results, called Laplacian Pyramid~\cite{Burt1983_LaplacianPyramid}, but it requires storing extra information in its algorithm, so it is not feasible to be applied as a reduced-order modelling framework.

\begin{figure}[!htb]
	\centering
	\includegraphics[width=\textwidth]{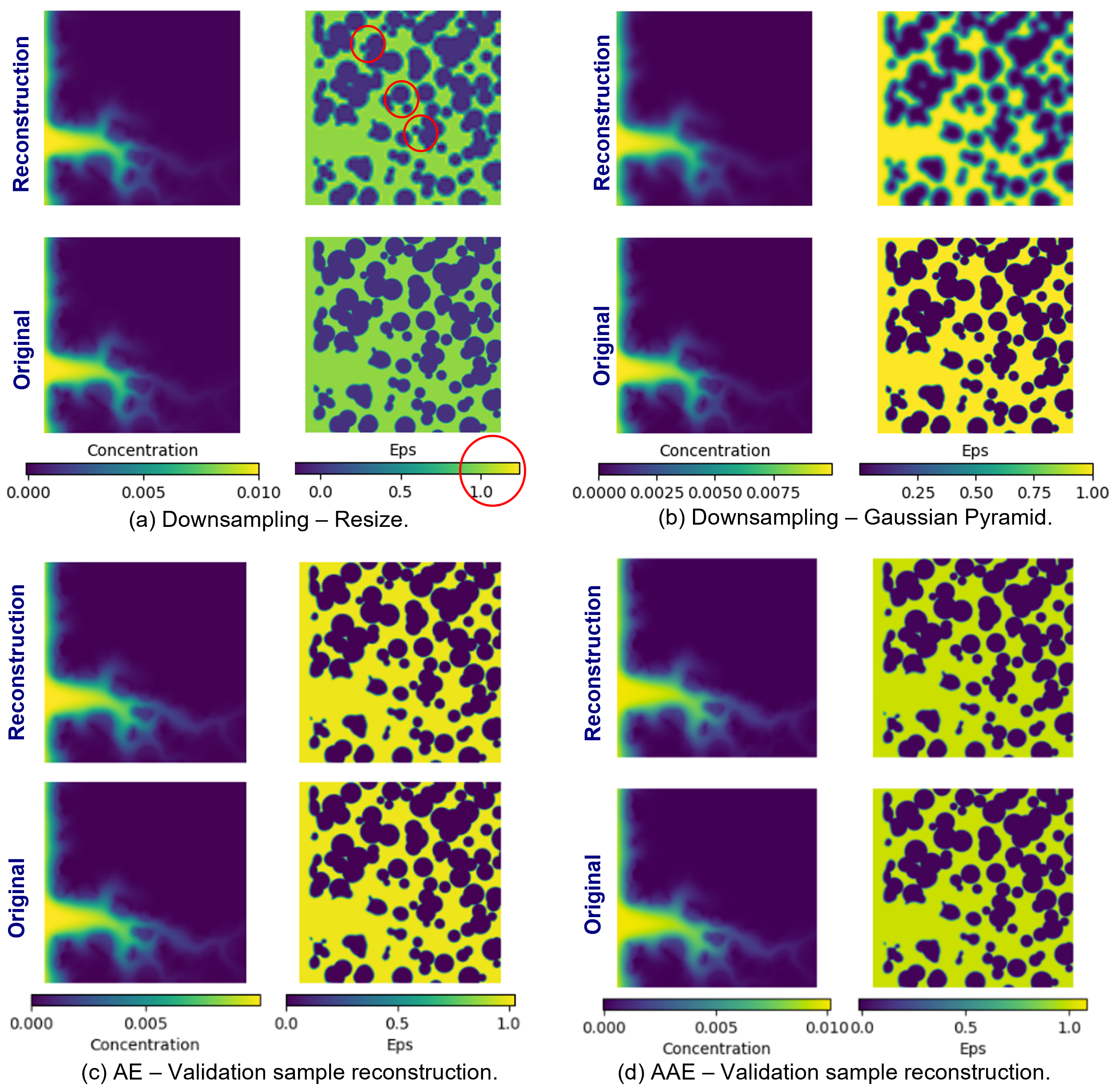}
	\caption{Comparison between downsampling methods common in image processing, the autoencoder used and the adversarial autoencoder used.}
	\label{fig:Compression-Results}
\end{figure}

The reconstructed output of the compression AE and AAE in a validating sample is nearly indistinguishable from the original fields, as can be seen in Figure \ref{fig:Compression-Results}. The average Mean Square Error obtained in the reconstruction of training data was \num{5.7e{-6}} and in validation data was \num{1.1e{-5}} for the autoencoder. For the adversarial autoencoder, it was \num{3.8e{-5}} in training data and \num{8.7e{-5}} in validation data.

\end{appendices}


\bibliography{bibliography}% common bib file

\end{document}